\documentclass[journal,twocolumn]{IEEEtran}

\usepackage{amsmath,amsfonts,amssymb,amsthm,mathtools}
\usepackage{comment,booktabs,hyperref,multirow,makecell}
\usepackage[dvipsnames]{xcolor}

\usepackage{algorithm,algpseudocode}

\usepackage{xcolor}

\newcommand{\ie}{i\/.\/e\/.,\/~}

\usepackage{cite}

\usepackage{fancyhdr}

\begin{document}
	\title{Multimodal Multi-User Surface Recognition \\with the Kernel Two-Sample Test}
	
	\author{Behnam~Khojasteh, \textit{Student Member}, \textit{IEEE}, Friedrich~Solowjow,\\ Sebastian~Trimpe, \textit{Member}, \textit{IEEE}, and~Katherine J. Kuchenbecker, \textit{Fellow}, \textit{IEEE}% <-this % stops a space
        	\thanks{B. Khojasteh and K.~J.~Kuchenbecker are with the Max Planck Institute for Intelligent Systems (MPI-IS), Stuttgart, Germany, and also with the Faculty of Engineering Design, Production Engineering and Automotive Engineering, University of Stuttgart, Stuttgart, Germany. F. Solowjow and S. Trimpe are with the Institute for Data Science in Mechanical Engineering, RWTH Aachen University, Aachen, Germany, and also with MPI-IS, Stuttgart, Germany. Email address for contact: khojasteh@is.mpg.de}
			\thanks{This work has been submitted to the IEEE for possible publication. Copyright may be transferred without notice, after which this version may no longer be accessible.}
			}
	
	\markboth{IEEE TRANSACTIONS ON AUTOMATION SCIENCE AND ENGINEERING,~Vol.~X, No.~X, MONTH~X}
	{Khojasteh \MakeLowercase{\textit{et al.}}: Multimodal Multi-User Surface Recognition with Kernel Two-Sample Test}
		
	\maketitle
		
\begin{abstract}  % max. 200 words; current 189 words
    Machine learning and deep learning have been used extensively to classify physical surfaces through images and time-series contact data.
    However, these methods rely on human expertise and entail the time-consuming processes of data and parameter tuning.  
    To overcome these challenges, we propose an easily implemented framework that can directly handle heterogeneous data sources for classification tasks.
    Our data-versus-data approach automatically quantifies distinctive differences in distributions in a high-dimensional space via kernel two-sample testing between two sets extracted from multimodal data (e.g., images, sounds, haptic signals).
    We demonstrate the effectiveness of our technique by benchmarking against expertly engineered classifiers for visual-audio-haptic surface recognition due to the industrial relevance, difficulty, and competitive baselines of this application; ablation studies confirm the utility of key components of our pipeline.
    As shown in our open-source code, we achieve 97.2\% accuracy on a standard multi-user dataset with 108 surface classes, outperforming the state-of-the-art machine-learning algorithm by 6\% on a more difficult version of the task.
    The fact that our classifier obtains this performance with minimal data processing in the standard algorithm setting reinforces the powerful nature of kernel methods for learning to recognize complex patterns.
    
\end{abstract}

\begin{abstract} % 100-300words ok; currently 190 words
    \textit{Note to Practitioners}: 
    We demonstrate how to apply the kernel two-sample test to a surface-recognition task, discuss opportunities for improvement, %\footnote{LINK}
    and explain how to use this framework for other classification problems with similar properties.
    Automating surface recognition could benefit both surface inspection and robot manipulation. 
    Our  algorithm quantifies class similarity and therefore outputs an ordered list of similar surfaces.
    This technique is well suited for quality assurance and documentation of newly received materials or newly manufactured parts.
    More generally, our automated classification pipeline can handle heterogeneous data sources including images and high-frequency time-series measurements of vibrations, forces and other physical signals.
    As our approach circumvents the time-consuming process of feature engineering, both experts and non-experts can use it to achieve high-accuracy classification.
    It is particularly appealing for new problems without existing models and heuristics. 
    In addition to strong theoretical properties, the algorithm is straightforward to use in practice since it requires only kernel evaluations.
    Its transparent architecture can provide fast insights into the given use case under different sensing combinations without costly optimization. 
    Practitioners can also use our procedure to obtain the minimum data-acquisition time for independent time-series samples from new sensor recordings.     
\end{abstract}


\IEEEpeerreviewmaketitle

\section{Introduction}
\IEEEPARstart{A}{utomation} expedites the capabilities of industrial systems by complementing and enhancing the skills of human workers. % functionalities
An emerging domain for automation is surface recognition: %, particularly being crucial for sequential manipulation.
identifying a surface through physical interaction is a challenging task that is highly relevant for manufacturing and construction. 
An artificial sensing system can gather and process multimodal data from human-guided interactions with physical objects to help inspect and identify their surfaces.
To emulate the richness of human perception, they need to evaluate diverse surface attributes \cite{richardson2020learning}.

\begin{figure}[tb]
	%	\centering
	\includegraphics[width=83mm]{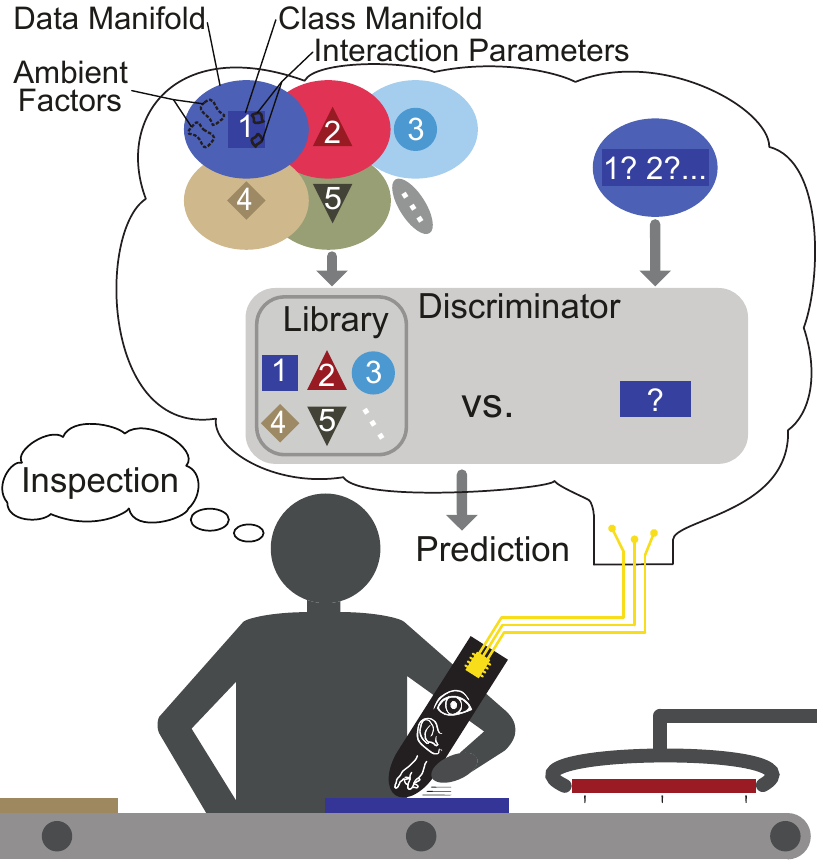}
	\centering
	\caption{Our general concept for the task of surface recognition. 
	Data manifolds that are generated from multimodal visual-audio-haptic information include both ambient factors (e.g., background noise) and complementary surface-specific subspaces that are spanned by interaction parameters (e.g., tool speed, normal force). 
	A well-performing discriminator retrieves the signature of an unlabeled inspection trial through comparison with a library of previously extracted class manifolds.}
	\label{fig:recognition_highlevel}	
\end{figure}

Multimodal sensor data from surface exploration span a rich high-dimensional space (Fig.~\ref{fig:recognition_highlevel}).
We are interested in identifying the surface from which unlabeled multimodal data were collected, recognizing that both the chosen interaction parameters and ambient factors strongly influence what is recorded.
For the modalities of vision, audio and touch, interaction parameters can include lighting, sensor location, scanning speed, applied normal force and the shape and properties of the sensing tool, while ambient factors might be temperature or background noise.
The major challenge of a generalizable surface recognition algorithm is to \textit{detect causal content-relevant properties inherent in multimodal information} so that these signatures can be \textit{retrieved from unseen data}.

How artificial systems can efficiently use multimodal information to infer surface properties is elusive.
% cross modal learning + all 3 citations use all LMT108 dataset
Recent advances in cross-modal learning algorithms for surface recognition include inferring images from tactile \cite{liu2018surface} or tactile-auditory cues \cite{zheng2019cross} and visual-to-tactile perception \cite{liu2020toward}.
Prior research demonstrated improved surface recognition when fusing multimodal sensor information~\cite{burka2016proton,strese2021haptic}.
However, it is more difficult to determine what combination of multimodal data is optimal for surface classification, particularly because performance depends on the algorithm employed.

As detailed in Section \ref{sec:literature}, machine learning and deep learning have both been used for surface classification. % passive
Traditional machine-learning classifiers rely on problem-dependent features to make predictions.
Determining suitable features is cumbersome and requires expert knowledge to search for content-relevant features, remove dependent features through correlation analysis, test feature quality to ensure intra- and inter-class uniqueness, and optimize features iteratively over the course of training. % statistical hypothesis testing
The ideal outcome of such a pipeline is an independent descriptive set of features that generalizes to unseen data.
However, the final results with the selected features strongly depend on subjective decisions, making it hard to gauge whether the given solution will suffer from overfitting.
Deep-learning algorithms operate without the need for feature engineering by efficiently learning patterns in the data that represent correlations within the imposed classes.
For instance, a deep-learning network trained only on images can achieve a high surface classification accuracy, but it may make critical misclassifications when applied to images only slightly outside the training distribution \cite{strese2019haptic,geirhos2020shortcut}. 
Multimodal data sources may facilitate more robust representations \cite{zheng2016deep,strese2021haptic}.
Apart from the ubiquitous issue of overfitting, deep-learning approaches require large amounts of data for training as well as significant effort and expertise to handle heterogeneous data sources and tune hyperparameters.
To conclude, there are many good reasons to automate the time-consuming processes of past approaches and try to learn directly from multimodal sensor data.

We implement a purely data-driven approach to multimodal classification by leveraging the kernel two-sample test of Gretton et al.~\cite{gretton2012kernel}. 
We build upon Solowjow et al.'s foundational use of this test to classify dynamical systems directly from their time-series outputs without feature engineering~\cite{solowjow2020kernel}. 
The main contributions of this paper are as follows:
\begin{itemize}
\item We improve and augment Solowjow et al.'s kernel-based classification framework \cite{solowjow2020kernel} to the task of multimodal multi-user surface recognition.
Our proposed approach can directly examine heterogeneous data sources such as images together with low- and high-frequency time series. 
\item We benchmark our algorithm on the task of surface recognition with Strese et al.'s multimodal dataset of $108$ surfaces~\cite{strese2017content}.
It outperforms the published results of expertly crafted machine-learning classifiers despite having a more difficult classification setting, minimal data processing, and minimal domain knowledge.
We publicly share the code of our surface-recognition pipeline to facilitate future comparisons.
\item Based on the surface-recognition task and backed up by ablation studies, we provide guidelines for using our efficient framework for other classification tasks.
The proposed method is readily applicable to various problems with heterogeneous data sources, spurious effects from temporal autocorrelations, scarce training data,  interaction parameters, and other ambient factors.
\end{itemize}

\section{Related Work}
\label{sec:literature}

Many algorithms have been proposed for object and surface recognition in the past decade.
The following summary divides this surface-classification research into the two common categories of expertly crafted features and deep-learned features.
This section concludes with a brief summary of prior research on our chosen tool for this task, the kernel two-sample test.

\subsection{Engineering Features with Machine-Learning Classifiers}

Human-subject studies of object and surface recognition tasks suggest that humans use a wide range of descriptions to categorize tactile sensations. % characterize
With regard to touch perception, four major tactile dimensions (hardness, roughness, warmth, and friction) have been identified \cite{holliins1993perceptual, tiest2006analysis}.
Further research resulted in refinement of the major tactile dimensions to five~\cite{okamoto2012psychophysical} and fifteen subdimensions~\cite{fishel2016method}. 

% tool-mediated freehand surface exploration
Strese et al.\ \cite{strese2016multimodal,strese2021haptic,strese2019haptic} conducted extensive research on feature-engineered surface classification and achieved impressive recognition accuracy for $69$ ($86$\%), $108$ ($91.2$\%) and $186$ ($90.2$\%) surfaces. % comprehensive
In the case of $69$ and $108$ surfaces, classification was performed with a training set of one expert and a testing set of ten other humans, while the training and testing data for the $184$ surfaces were acquired by the same person; generalizing to surface-interaction data from other humans is known to be more difficult \cite{shao2016spatial,strese2016multimodal,strese2021haptic,liu2022texture}.
Strese et al.\ used traditional machine-learning classifiers such as k-nearest neighbor (k-NN), random forests, or combinations of multiple classifiers.
Their features stem from auditory, visual, and haptic data as well as data traces outside human sensing capabilities, e.g., infrared (IR) light reflectance and electrical conductivity.
Based on their most recent material-scanning systems, they propose $15$ perceptual features for multimodal data \cite{strese2019haptic}.
Another study by Burka et al.\ \cite{burka2017handling} boosted the classification accuracy on the dataset of $69$ surfaces up to $87.5$\% using a multi-class support vector machine (SVM) classifier with the same hand-engineered features. 
For new surface recordings, Burka et al.\ also demonstrated superior classification performance of Strese et al.'s features over simpler scan-dependent features, which change with applied force and tool tip speed \cite{burka2017handling}.
Lu et al.\ \cite{liu2022texture} recently achieved $74\%$ classification accuracy with multi-user auditory-haptic data for nine texture classes on the $69$-surface dataset using a novel spectral feature; their Naive Bayes classifier used the data from the ten participants in a $7$ vs.\ $3$ train-test split.

% robotic scanning systems
Several studies of robotic systems with planar scanning showed high surface-classification accuracies. 
For instance, Sinapov et al.\ \cite{sinapov2011vibrotactile} used a k-NN classifier with spectrotemporal acceleration features, and Jamali et al.\ \cite{jamali2011majority} presented a Naive Bayes classifier with temporal mean as well as distinctive frequency features from strain gauge readings. 
Fishel and Loeb \cite{fishel2012bayesian} calculated statistical features from multiple robot-controlled scans by a biomimetic multi-stream tactile sensor (SynTouch BioTac) and used a Naive Bayes classifier to achieve $95.4$\% recognition accuracy among $117$ textures.

In the field of robotic interaction with three-dimensional objects, Chu et al.\ \cite{chu2015robotic} showed that a robot equipped with two BioTacs can achieve comparable performance to average humans at labeling objects with $25$ haptic adjectives.
They implemented a multi-kernel SVM classifier that extracted standard static features and dynamic features given by the first four statistical moments of spectral vibrations. 
Kaboli and Cheng \cite{kaboli2018robust} extended this idea with a robotic hand with five BioTacs and successfully classified $30$ in-hand objects ($98$\%) and $120$ surfaces ($100$\%).

\subsection{Learning Features with Deep-Learning Classifiers}

An alternative approach for learning surface classes consists of training neural networks that directly learn inherent features in a latent space. %
Such neural networks generally contain input and output layers as well as hidden layers that can all interact through nonlinear mathematical operations.
Deep-learning architectures are known to be powerful in the presence of non-stationary textures and varying measurement conditions~\cite{bello2019comparative}. 

% vision
Convolutional neural networks have been successfully used for image-based recognition of objects \cite{krizhevsky2012imagenet}. 
A review paper \cite{liu2019bow} summarizes that advanced network structures like AlexNet, VGGM, VGGVD, and TCNN were able to accurately classify $95$ to $98$\% \cite{cimpoi2015deep} of the largest existing visual texture databases (CUReT, KTH-TIPS2, ALOT).  % omitted dataset and network references 

% multimodal
Going beyond purely vision-based approaches, Zheng et al.\ \cite{zheng2016deep} fuse tool-contact accelerations and images that were acquired by one expert in a fully convolutional network that can classify $69$ surfaces with $98\%$ accuracy. 
Particularly, their deep-learning classifier extracts time-frequency information from acceleration signals in order to learn latent features from the tactile domain.
Joolee et al.\ \cite{joolee2022deep} addressed the same classification setting with accelerations alone; their multi-model fusion network achieved $98.18\%$ accuracy, outperforming several machine- and deep-learning methods.
Gao et al.\ \cite{gao2016deep} applied hybrid convolutional and recurrent networks to new object images and Chu et al.'s previously recorded haptic data from exploratory procedures on $53$ objects \cite{chu2015robotic}; this learning-focused approach outperformed machine-learning classifiers based on hand-designed features.
Strese et al.\ \cite{strese2019haptic} slightly outperformed ($90.7$\% vs. $90.2$\%) their own machine-learning classifier, which incorporated data from auditory, haptic, and visual modalities, by considering only images with a modified convolutional network from AlexNet \cite{krizhevsky2012imagenet}. 
However, their superior image-based deep neural network exhibited more critical misclassifications for the $184$ surfaces due to its lack of information about the contact properties. 
The reasonableness of mistakes is an important way to judge classification techniques to be used in real applications.
Wei et al.\ \cite{wei2021multimodal} recently used solely auditory-haptic data to  achieve $87.6\%$ classification accuracy for eight material categories. %from introduced taxonomy \cite{strese2019haptic}.
The architecture of their multimodal convolutional neural network efficiently used multi-scale temporal information for classifying new surfaces.

\subsection{Learning with the Kernel Two-Sample Test}

% %%%
Kernel methods that operate in a high-dimensional space are a powerful and popular tool in machine learning \cite{scholkopf2018learning}. 
In addition to mapping single data points, as used for SVMs, one can also directly map full distributions into reproducing kernel Hilbert spaces (RKHS).
Due to its structure, the Hilbert space is equipped with a scalar product, and scalar products can be evaluated through kernel evaluations.
The embedding of probability distributions is called kernel mean embeddings; a recent review by Muandet et al.\ \cite{muandet2017kernel} provides an extensive summary of existing work.
In particular, kernel mean embeddings yield tractable approximations for the maximum mean discrepancy (MMD), a metric on the space of probability distributions.
The MMD metric quantifies the difference between two distributions by taking into account not only low-order statistical moments (e.g., mean, standard deviation, skewness, kurtosis) but also higher-order statistical moments.
The MMD can be well suited as a discriminatory metric for classification tasks, when data distributions between imposed classes differ.
Gretton et al.\ \cite{gretton2012kernel} derived a rigorous MMD-based statistical test: at its core, it estimates the MMD between two datasets and tests the null hypothesis that they were drawn from the same distribution.
They developed this so-called kernel two-sample test for both numerical examples and real-world applications.
Since then, the kernel two-sample test has emerged as a prominent method in probabilistic machine learning due to its powerful theoretical guarantees and versatile applicability~\cite{long2015learning,muandet2017kernel,TolBouGelSch18,rozantsev2018beyond,vankadara2021recovery}.

Several extensions of the kernel two-sample test have been developed \cite{doran2014permutation,chwialkowski2014kernel,lloyd2015statistical}.
In particular, Lu et al.~\cite{lu2022preference} recently used the MMD metric to quantify the similarity between real and synthesized surface signals.
Further, satisfying the assumption of independent and identically distributed (i.i.d.)\ data for the kernel two-sample test can be particularly difficult for certain applications.
For the surface-recognition task, we will work with non-i.i.d.\ time-series data.
Therefore, we will leverage recent results from prior work \cite{solowjow2020kernel} that extends the kernel two-sample test to dynamical systems with non-i.i.d.\ data.
In addition to the theoretical foundation, Solowjow et al.\ also show the efficiency of their framework on a gait classification task with two labels using low-frequency periodic IMU signals.
This novel approach of comparing dynamical systems is relevant in the context of surface interactions, particularly because the MMD can consider multiple data points from the evolution of a time series.

\section{Problem Formulation}

%%%% % High-level ‘intuitive’
Our goal is to classify unseen multimodal sensor recordings from physical surface interactions. % with a surface.
The algorithm for this task needs to process heterogeneous and potentially high-dimensional data (Fig.~\ref{fig:recognition_highlevel}).
In particular, it needs to retrieve the surface signature of an unlabeled data manifold from a library of representative classes.
Since these meaningful signatures are hard to obtain by applying first principles or training models, a more pragmatic approach is necessary.

%%%% % Low-level ‘mathematical'
Mathematically, we approach the surface-recognition task purely by comparing data distributions. 
We model surface interactions as realizations of a stationary stochastic process or dynamical system, similar to prior work by Solowjow et al.~\cite{solowjow2020kernel}. 
We assume that a set of $C\in \mathbb{N}$ unique surfaces  will induce different distributions $\mathbb{P}_1,\ldots,\mathbb{P}_C$, respectively.
Our goal is to assign unlabeled surface recordings made by a new human user to the corresponding surface distribution to determine the surface class $c$ from which it most likely came.

\begin{figure}[thb]
	%	\centering
	\includegraphics[width=\columnwidth]{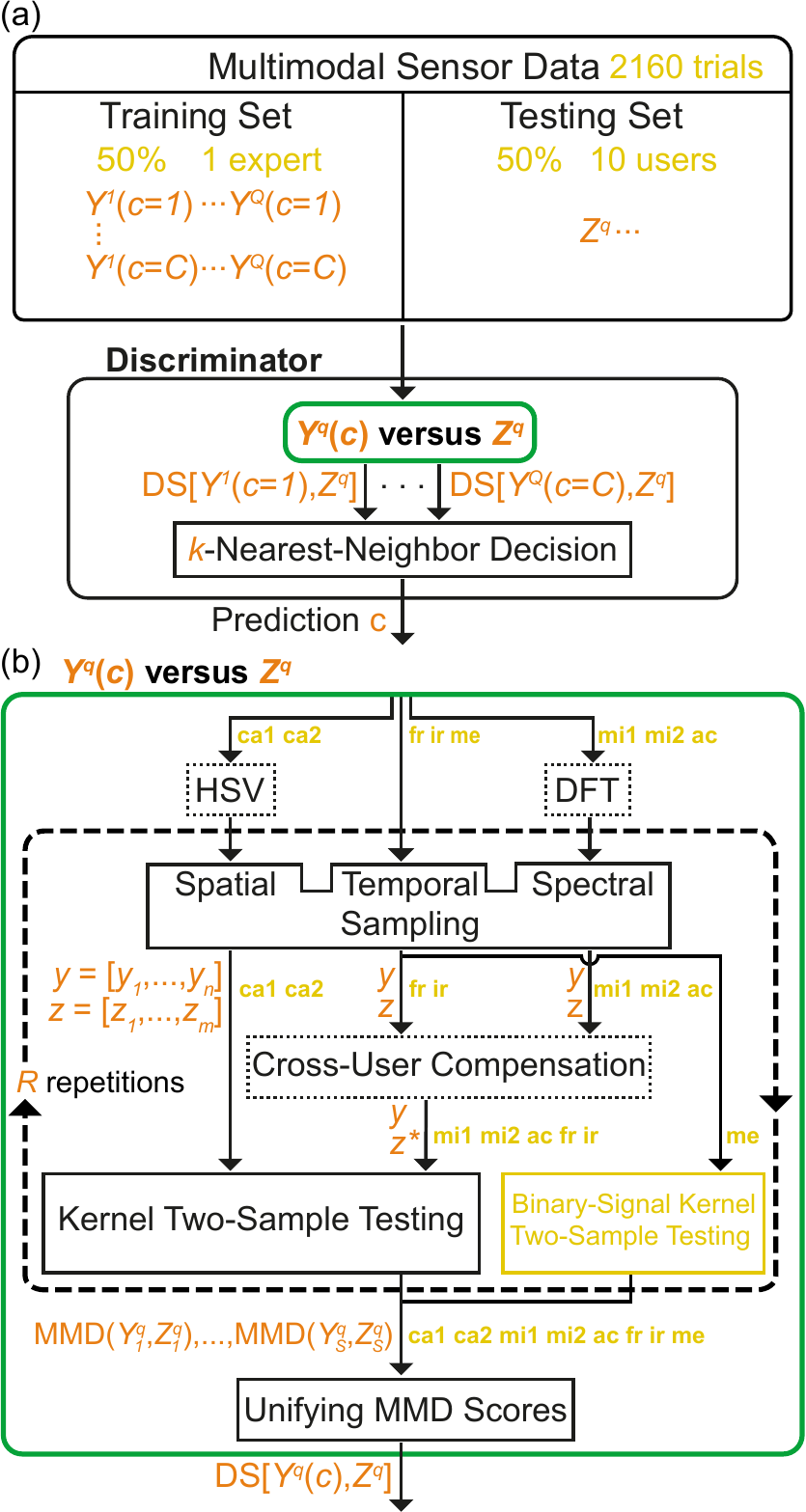}
	\centering
	\caption{Proposed surface-recognition pipeline that directly handles data without feature engineering. 
	Orange highlights the introduced nomenclature and yellow the dataset-specific information.
	(a) Our discriminator processes multimodal data to make a surface-identity prediction.
	The training set consists of a surface data library from one user, and the testing set contains data from ten other users.
	(b) At the core of our approach, the discriminator compares two data trials, $Y^q(c)$ and $Z^q$, across $S$ information sources.
	To elucidate the effectiveness of the recognition pipeline, ablation studies are conducted for the three dotted function blocks (HSV, DFT and cross-user compensation).} 
	\label{fig:pipeline}
\end{figure}

\section{Proposed Method}

We propose a recognition framework (Fig.~\ref{fig:pipeline}(a)) to solve the aforementioned classification problem. 
We assume there are $Q \in \mathbb{N}$ trials per surface class $c$, where each trial consists of data from $S\in \mathbb{N}$ information sources.
Consequently, we obtain for the $q$-th surface trial $Y^{q}=\{Y^{q}_1,\ldots,Y^{q}_S\}$ a set of information sources. 
Information sources can include data from different sensors and/or different exploration procedures with the same sensor.
In our classification setting, the surface trials $Y^q$ and $Z^q$ belong to the training and testing sets, respectively.
At the core of our approach (Fig.~\ref{fig:pipeline}(b)), we extract two sample sets, $[y_1,\ldots,y_n]$ and $[z_1,\ldots,z_m]$ with corresponding sample sizes $n$ and $m$, from two data streams, $Y^{q}_s$ and $Z^{q}_s$, with unknown distributions, $\mathbb{P}_{Y^{q}_s}$ and $\mathbb{P}_{Z^{q}_s}$.
We then quantify the similarity in distribution of the two sample sets in the given information source $s$.
A discrepancy score (DS), \ie $\mathrm{DS}[Y^q,Z^q]$, unifies the computed similarity values across all $S$ information sources.
By pairwise comparisons with all data in a class library, we assign the trial $Z^{q}$ to a distribution and thus to a surface class $c$. 

For the surface-recognition task, we consider a public tool-surface dataset \cite{strese2017content} that consists of images and time-series data recorded by eleven individuals.
Our recognition pipeline (Fig.~\ref{fig:pipeline})  relies solely on an offline data library of measurements by one expert; no feature extraction is needed during this passive training phase.
In the testing phase, we compare the data distributions of each surface in the library with unlabeled measurements from the ten other users, \ie one expert vs.\ ten users. 
Our framework accommodates heterogeneous input sources, deals with temporal autocorrelations, and compensates for user- and session-dependent effects in order to make a good classification decision.
The following subsections detail the functionality of our recognition paradigm with the kernel two-sample test.
Appendix~\ref{sec_softwareImplements} provides links to the dataset and the code.

\subsection{A Metric on the Space of Probability Distributions}

We briefly introduce the technical concepts necessary for this article; for details we refer to the review paper by Muandet et al.\ \cite{muandet2017kernel}.
We use kernel mean embeddings to quantify the difference between two data distributions.
Kernel mean embeddings lift distributions into a higher and potentially infinite-dimensional RKHS, where certain tasks become tractable. 
Under some mild assumptions on the kernel $k$, it can be shown that the embedding is injective, 
which ensures that only one object is mapped to zero; this property is the foundation for meaningful statistical tests~\cite{muandet2017kernel}. These kernels are called characteristic. 
Let $\mathcal{H}$ be an RKHS, $\mathcal{Y}$ the data input space, and $\phi : \mathcal{Y} \rightarrow \mathcal{H}$  its canonical feature map  for data points, $y_i,z_i\in \mathcal{Y}$, so that $\phi(y),\phi(z)\in \mathcal{H}$.
As the mapping $\phi$ to the RKHS is usually unknown,  the so-called kernel trick enables the replacement of the inner products by evaluations of a kernel $k$,
so that
\begin{equation}\label{kerneltrick}
\langle\phi(y),\phi(z)\rangle_{\mathcal{H}} =: k(y,z).
\end{equation}

The kernel two-sample test \cite{gretton2012kernel} is designed to distinguish between probability distributions, $\mathbb{P}_Y$ and $\mathbb{P}_Z$, based on i.i.d.\ samples from these distributions. 
At its heart, the kernel two-sample test estimates the MMD, which is defined as 
\begin{equation}
\label{eq:MMDana}
\begin{split}
    \mathrm{MMD}(\mathbb{P}_Y, \mathbb{P}_Z,\mathcal{H}) &= \sup_{ \lVert f\rVert_\mathcal{H} \leq 1}    \mathbb{E}_{\mathbb{P}_Y}[f(y)] - \mathbb{E}_{\mathbb{P}_Z}[f(z)]   \\
    &= \| \mu_{\mathbb{P}_Y} - \mu_{\mathbb{P}_Z}\|_\mathcal{H}
    % \Biggl\{ \Biggr\}
    \end{split}
\end{equation}  
and is a metric on the space of probability distributions.
To compare distributions, the MMD considers functions $f$ in the unit ball of the RKHS $\mathcal{H}$ induced by the kernel $k$.
Equivalently, we can express the MMD in terms of the kernel mean embeddings $\mu_{\mathbb{P}_Y}$ and $\mu_{\mathbb{P}_Z}$, which yields the second equality (see \cite{muandet2017kernel} for details).
A kernel mean embedding is a powerful statistical measure as it contains all statistical moments of a distribution. 

In general, it is intractable to compute the MMD in Eq.~\eqref{eq:MMDana} directly. 
However, there are kernel-based estimates that efficiently approximate the MMD.
Furthermore, the convergence speed can be quantified based on concentration results. 
Thus, it is possible to construct high confidence intervals and a significance-level-$\alpha$ statistical test.
We use the squared MMD estimator by Gretton et al.\ \cite{gretton2012kernel}, 
\begin{equation}\label{eqn:mmdComputable}
\begin{split}
	\mathrm{MMD}^{2}_b[\mathbb{P}_Y,\mathbb{P}_Z] = \dfrac{1}{n^2} \sum_{i,j=1}^{n} k(y_i,y_j)  \\
	+  \dfrac{1}{m^2}\sum_{i,j=1}^{m} k(z_i,z_j) - \dfrac{2}{nm} \sum_{i=1}^{n}\sum_{j=1}^{m} k(y_i,z_j),
	\end{split}
\end{equation}
where 
$[y_1,...,y_n]$ and $[z_1,...,z_m]$ are i.i.d.\ random variables.
In our case, these are $n$ and $m$ samples from surface data streams $Y^{q}_s$ and $Z^{q}_s$ with unknown distributions $\mathbb{P}_{Y^{q}_s}$ and $\mathbb{P}_{Z^{q}_s}$. 
The two-sample test maximizes over an infinite-dimensional feature space of possible low- and high-order statistics and automatically selects the statistical properties with highest discrepancy. 
No prior knowledge or parameterization about $\mathbb{P}_{Y^{q}_s}$ and $\mathbb{P}_{Z^{q}_s}$ is required to obtain the difference between the distributions. 

In practice, an estimate of the MMD will fluctuate even for identical distributions due to the finite number of sampled data points.
A threshold $\kappa$ for the two-sample test ensures confidence bounds with significance level $\alpha$ for the empirical MMD under the null hypothesis that the two distributions are not different. 
If the empirical squared MMD in Eq.~\eqref{eqn:mmdComputable} fulfills  
\begin{equation}\label{kappa_test}
\mathrm{MMD}^{2}_b[\mathbb{P}_{Y^{q}_s},\mathbb{P}_{Z^{q}_s}] > \kappa(n,\alpha),
\end{equation}
one can conclude that $\mathbb{P}_Y \neq \mathbb{P}_Z$ with high probability $1 - \alpha$.

The choice of the kernel function $k$ can be critical, and there are many valid choices with potential hyperparameters that can be optimized.
For this paper, we use the prominent candidate, the squared exponential function,
\begin{equation}\label{gaussKernel}
%	k(a_i,b_j) = \exp(-\dfrac{ || a_i - b_j||^2 }{2\sigma^2})
	k(y,z) = \exp\left(-\dfrac{ \| y - z\|^2 }{2\sigma^2} \right),
\end{equation}
as the kernel function for all statistical tests.
For every choice of the length scale $\sigma\in \mathbb{R}^+$, we obtain a valid kernel function. 
However, the performance of the algorithm will generally depend on the choice of the hyperparameter $\sigma$, since we consider a finite amount of data. 
Instead of optimizing over  $\sigma$, we choose the well-established median heuristic \cite{gretton2012kernel}  
\begin{equation}\label{medianHeuristic}
	\sigma^2 = 0.5 \cdot\mathrm{median}(\{\| z_i-z_j\|^2\}) 
\end{equation}
over distances between a subset of samples ($i,j=1,\ldots,100$).

\subsection{Sampling from Time-Series Data} 

As the significance test in Eq.~\eqref{kappa_test} requires independent data, we need to address inherent autocorrelations before using time-series data for kernel two-sample tests.
Autocorrelations in time series are one type of serial dependence that represent temporally correlated observations.
Coping with autocorrelations in dynamical systems or stochastic processes is challenging, but it is critical to preserve theoretical guarantees for time-series analysis.
High-order pattern recognition tasks particularly rely on removing spurious autocorrelations through, e.g., mutual information, as shown in prior work on cutaneous friction patterns \cite{khojasteh2018complexity}.

Mutual information and Pearson's correlation are common techniques to measure the dependence between two variables.
Here, we follow the MMD-compatible approach from prior work \cite{solowjow2020kernel} and use the Hilbert-Schmidt independence criterion (HSIC) to quantify dependence.
The HSIC between two random variables, $Y$ and $Z$, can be defined \cite{sejdinovic2013equivalence} as
\begin{equation}
\label{eq:HSICana}
    \mathrm{HSIC}(Y,Z) = \|\mathbb{P}_{Y} \otimes \mathbb{P}_{Z} -  \mathbb{P}_{Y,Z}\|_{\mathrm{MMD}},
\end{equation}
with the tensor product $\otimes$ of the two marginal distributions, $\mathbb{P}_{Y}$ and $\mathbb{P}_{Z}$, and their joint distribution $\mathbb{P}_{Y,Z}$ under the MMD metric.
It is generally intractable to compute HSIC as defined in Eq.~\eqref{eq:HSICana}.
Thus, we use a kernelized estimate \cite[Eq. (4)]{gretton2007kernel} 
\begin{equation}\label{eqn_hsicMatrix}
\mathrm{HSIC}_b(Y,Z) = \dfrac{1}{n^2} \mathrm{tr}(KHLH)
\end{equation}
with $H, K, L \in \mathbb{R}^{n\times n}$ and kernel operators $K_{i,j} = k(y_i,y_j)$ and $L_{i,j} = l(z_i,z_j)$. 
The centering matrix
\begin{equation}\label{eqn_hsicCentering}
H= I - n^{-1}ee^\top
\end{equation}
results from the identity matrix $I$ and an all-ones vector $e$.

In our setting, the time-series measurements represent observations from a physical interaction with a surface.
We assume that each trial $Y^{q}_s$ with $q=1,\ldots,Q$ of a given time-series source $s$ generated from interaction with surface $c$ is an independent process realization of the same dynamical system.
We extract two $t$-spaced data points, \ie $Y^q_s(t_1)$ and $Y^q_s(t_1+t)$, from each of the $Q$ realizations  and divide the data into two sets $Y:=  \{Y^1_{s}(t_1),\ldots,Y^Q_{s}(t_1)\}$ and $Z:= \{Y^1_{s}(t_1+t),\ldots,Y^Q_{s}(t_1+t)\}$ that are, by construction, i.i.d.\ within themselves. % (e.g., surface label)
Thus, $n$ from Eq.~\eqref{eqn_hsicMatrix} equals $Q$.
If we choose the time gap $t$ between consecutive data points to be large enough, the empirical estimator $\mathrm{HSIC}_b$ predicts approximately independent data, which is again consistent with the setting in Eq.~\eqref{kappa_test}.

In practice, the computation of approximately independent time-series samples takes several steps; see Algorithm \ref{algo_hsic}. 
At the core of the procedure, we extract from all $Q$ realizations $t$-spaced data points, $Y$ and $Z$, to compute $\mathrm{HSIC}_b$ and the threshold $\kappa$.
The threshold results from a bootstrapping approach with significance level $\alpha$.
Based on the application, the parameter $\alpha$ can be modified to control the test error during statistical testing; for instance, $\alpha=0.05$ limits the false positive error rate to $5\%$.
The bootstrapping procedure (see Appendix \ref{sec_appendixMIXproperties} for implementation details)  consists of computing the $\mathrm{HSIC}_b$ repeatedly with shuffled matrix $L$ and inferring $\kappa$ from the ($1-\alpha$)-quantile.
We repeatedly ($R$ times) perform this core procedure with randomized starting time $t_1$ to increase the accuracy of the statistical test. 
By performing the aforementioned steps with increasing time gap $t$, we will eventually obtain approximately independent data. % for suitable processes.
Our algorithm then outputs an optimal time separation $T^*$ that ensures that the upper $95\%$ confidence interval bound of the $\mathrm{HSIC}$ test power ($CB^+$) falls under the averaged test threshold $\overline{\kappa}$ for all $R$ repetitions. % (see examples for two different surfaces Fig.~\ref{fig:hsic}).
Thus, we can reliably guarantee that autocorrelations vanish in the course of time-series data extraction for the MMD test.
The choice for upper initialization range $t_{1,\max}$ bounds the start of extracting sub-trajectories, which is negligible for stationary processes; however, together with the overall time-series duration $T_{\mathrm{total}}$, the parameter $t_{1,\max}$ determines the upper search bound for the optimal $T^*$ through $T \leq T_{\mathrm{total}} - t_{1,\max}$.

\begin{algorithm}[t] 
\caption{Minimum Distance for Independent Samples}
\label{algo_hsic}
\begin{algorithmic}[1]
\Require{$Y^{1,...,Q}_s$, $R$, $t_{1,\max}$, $\alpha$} 
\Ensure{Vanishing autocorrelation time \textbf{$T^*$}}
\State Initialize $t=0$ and $CB^+>\overline{\kappa}$
\While{$CB^+>\overline{\kappa}$} % (CI(testStatistic) > $\kappa_{mean}$))            
    \State $t = t+1$
    \For{$r \gets 1$ to $R$}        
        \State Randomly initialize $t_1$ from $[1,t_{1,\max}]$
        \State Construct $Y$ and $Z$ with $t_1$ and $t$
        \State Compute $\mathrm{HSIC}_b(Y,Z)^r$ with Eq.~\eqref{eqn_hsicMatrix}
        \State Compute $\kappa^r$ by $\alpha$-bootstrapped $\mathrm{HSIC}_b(Y,Z)^r$  
    \EndFor
    \State {$\overline{\kappa}$ $\gets$ Average $\kappa^{1,...,R}$}
%    \State  {$CB^+$ $\gets$ $95\%$ t-statistics of $HSIC_b(Y,Z)^{1,...,R}$}
    \State  {$CB^+$ $\gets$ $(1-\alpha)$-quantile of $\mathrm{HSIC}_b(Y,Z)^{1,...,R}$}
\EndWhile
\State \Return $T^*=t$     %{$T^*= t$}
\end{algorithmic}
\end{algorithm}

The vanishing autocorrelation time $T^*$ indicates the mixing speed of temporal dependencies and gives insight into the complexity of the observed dynamical system.
Therefore, these mixing properties express the predictability of the process underlying the time series, analogous to entropy-related mutual information.
The speed of mixing is further elaborated and applied to sample surface signals in Appendix \ref{sec_appendixMIXproperties}.

\begin{figure*}[tb]
	%	\centering
	\includegraphics[width=170mm]{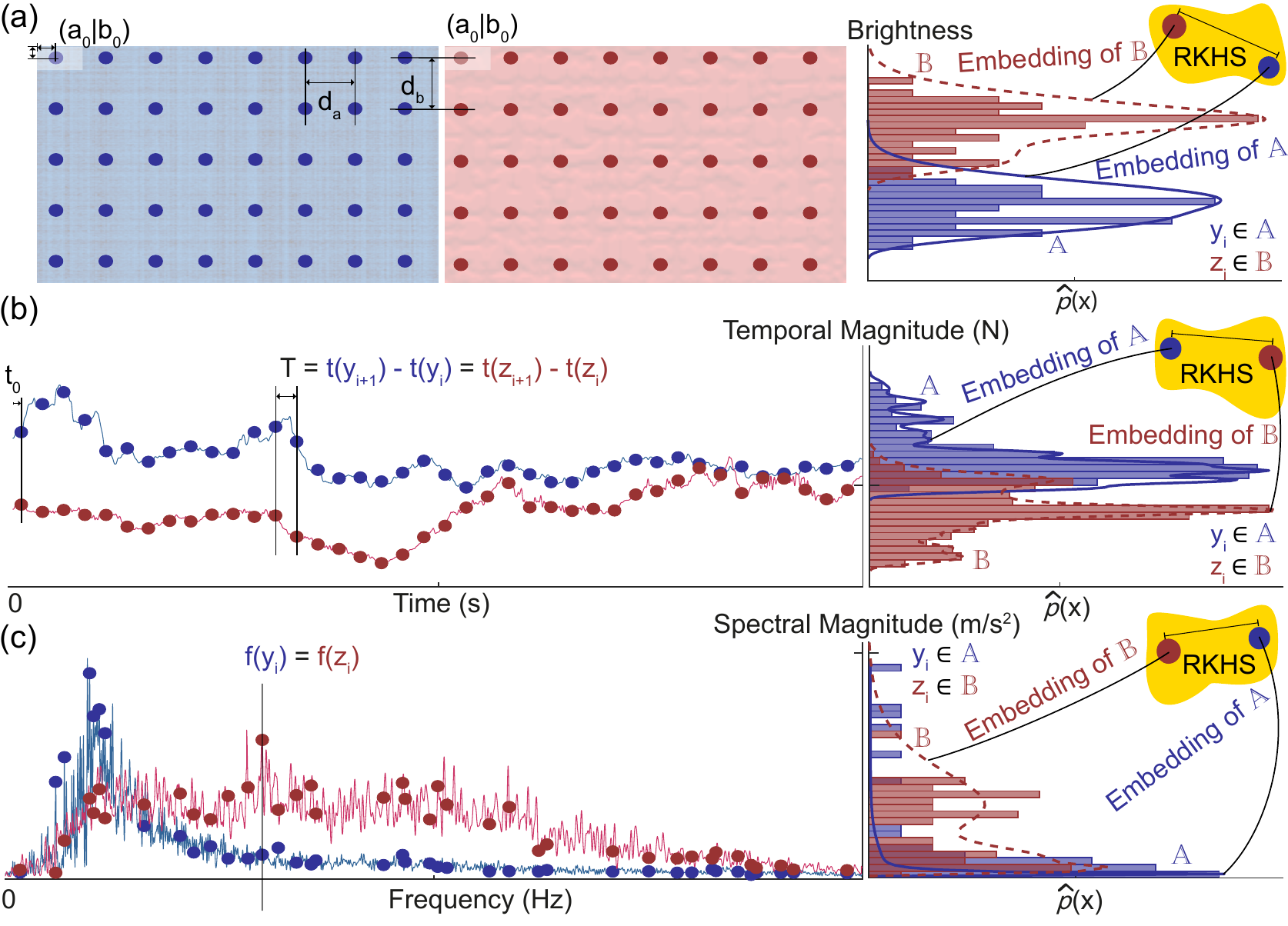}
	\centering
	\caption{Illustration of our sampling strategy for the kernel two-sample test in the (a) spatial, (b) temporal and (c) spectral domains. 
	While the sampling occurs in an equidistant fashion for the image and temporal signal values, a uniformly distributed  random, unique sampling scheme finds use in the frequency domain. 
	We extract spectral magnitudes at the same frequency locations for the two signals being compared.
	The kernel two-sample test  embeds the extracted sample points with their underlying distribution into a high-dimensional space of inner products, the RKHS.
	The MMD metric ensures maximization of the distinguishability of the two distributions in an automated fashion.}
		\label{fig:samplingStrategy}
\end{figure*}

\subsection{Sampling Strategy} % Sampling procedure

The kernel two-sample test benefits from unprocessed data channels  to maximize discrepancy and is also known to be computationally efficient on high-dimensional data \cite{gretton2012kernel}. 
Therefore, we primarily input raw sensor readings into the test and refrain from data-processing steps such as filtering or dimensional reduction (e.g., multichannel signal compression).

%%How to sample from heterogeneous, multimodal data for our statistical test.
In addition to classical time series, one can also input other data sources into the kernel two-sample test.
For example, spatially distributed pixel values of images and Fourier-transformed data can be highly expressive and have been used for decades in signal-processing applications.
Frequency-domain data in particular exhibits stochastic independence and therefore is naturally compatible with the MMD test.
Different color representations such as RGB (red, green, blue) and HSV (hue, saturation, value) have been widely used in the computer-vision community for the analysis of images; HSV in particular encapsulates important image properties such as pure color and its brightness value in separate channels. 

Conceptually, one could input the data streams $Y^{q}_s$ and $Z^{q}_s$ in both time and frequency domains into the kernel two-sample test, albeit at a higher computational cost.
However, for certain applications data might be sufficiently expressive in one of the two domains. 
We propose to choose the sampling domain based on whether meaningful surface information is encoded in the mean value of a data source, \ie the DC component of the signal. 
For instance, the average pixel intensity in an image may be characteristic of surface identity, whereas the high-frequency waveform of zero-mean data sources such as auditory and haptic vibrations is more distinctive.
We use this criterion to propose equidistant and random sampling strategies (Fig.~\ref{fig:samplingStrategy}) to extract data points for the MMD test. 

The equidistant sampling addresses images and time series with a non-zero signal mean and occurs in a spatial or temporal manner (Fig.~\ref{fig:samplingStrategy}(a) and (b)).
We randomly select the first data point at location ($a_0$, $b_0$) or $t_0$ within an initial image area or initial time window.
For simplicity, the location of the first sample ($y_1$ and $z_1$) can be the same for the two data streams. % For instance, the location...
We then equidistantly extract new data points with spatial or temporal distance, $d_a$ and $d_b$ or $T$, respectively, to cover the data as broadly as possible.
This sampling strategy results in a grid for images.
For time series, the distance between two sequential data points should be greater than or equal to $T^*$ in order to avoid autocorrelations.

For time series with high-frequency characteristics, we perform the random sampling scheme in the frequency domain (Fig.~\ref{fig:samplingStrategy}(c)).
This approach is well-suited for data with class-specific frequency spectra that will differ quantitatively in a classification context.
For example, accelerometer readings from tool-surface interaction exhibit texture-specific spectral characteristics \cite{culbertson2013generating,strese2016multimodal}. 
We convert the signal streams into the frequency domain by means of the discrete Fourier transform (DFT). 
We then randomly extract a set of absolute spectral magnitudes at the same, unique frequency locations for both signal streams. 
Thus, we obtain a direct comparison of the presence of these frequencies in both datasets.

The computational complexity of the kernel two-sample test nonlinearly increases with the number of extracted data points $n$ and $m$ in Eq.~\eqref{eqn:mmdComputable}. 
To reduce the computation time while maintaining a high classification accuracy, we repeatedly execute the MMD test with a reduced number for $n$ and $m$. 
The input data points across repetitions of the MMD test can differ in initial sample and/or inter-sample spacing.
Due to the law of large numbers, the average of the individual MMD estimates converges to the actual MMD. 
Thus, we can process data more efficiently, which leads to a higher accuracy for our use case.

\subsection{Compensation for User Differences}
Sensor measurements can vary significantly across different humans for the same task. 
This user variability is for instance observed in tool-surface vibrations \cite{culbertson2013generating} and cutaneous sliding friction forces from simple textured surfaces \cite{janko2015frictional}.
Causes of such cross-user differences can include the scanning speed, applied touch force, and the orientation of the sensing tool relative to the surface.
It is difficult to mitigate cross-user effects in signals such as contact forces for user-invariant pattern analysis even with a controlled measurement protocol, advanced signal-processing \cite{janko2015frictional}, and higher-order nonlinear techniques \cite{khojasteh2018complexity}. 

To illustrate that our approach for surface classification generalizes to many users, we need to mitigate the effect of the human. 
We pursue a holistic, statistical approach to compensate for individual differences; this approach is particularly advantageous when a dataset lacks relevant causal interaction parameters to model user differences. 
In order to standardize data between different users for the kernel two-sample test, we shift the mean $\overline{z}$ of the extracted data points $z=[z_1,\ldots,z_m]$ from the unlabeled data stream $Z^{q}_s$
\begin{equation}\label{eqn:scalShift}
z^* =  z + (\overline{y} - \overline{z})\cdot e  %\dfrac{\sigma_z}{\sigma_y}
\end{equation}
to the mean $\overline{y}$ of  $[y_1,\ldots,y_n]$ from data stream $Y^{q}_s$ in the library using an all-ones vector $e$.
The idea behind this cross-user compensation trick is that the effect of certain interaction parameters and ambient factors are essentially concentrated in the distribution mean and not in high-order moments, which ideally also contain information about surface properties.

As the user greatly influences contact-based measurements through their motion and other dynamical factors, we apply the cross-user compensation to all information sources that are sensitive to these dynamical user effects. 
For instance, changes in interaction parameters such as applied force or scanning speed between users might cause different friction force levels or vibrational spectral energy. 
Depending on the sampling domain, this shift operation standardizes the temporal signal mean or its spectral energy between data distributions from two users. 
As a result of the mean-aligned distributions, the data stream comparison focuses only on the higher-order statistical moments. 

\begin{figure*}[bht]
	%	\centering
	\includegraphics[width=170mm]{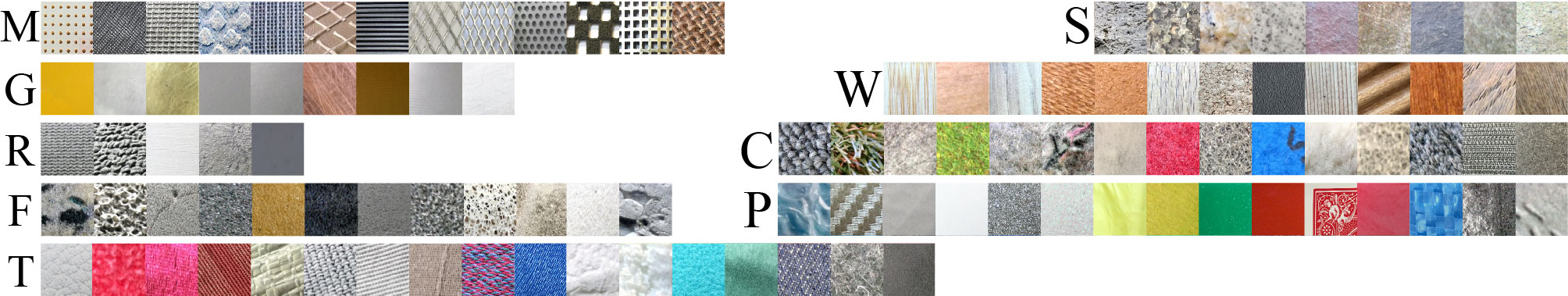}
	\centering
	\caption{The online LMT108 dataset \cite{strese2017content} covers 108 surfaces from nine material categories: (M) thirteen meshes, (S) nine stones, (G) nine glossy materials, (W)  thirteen wooden materials, (R)  five rubbers, (C)  fifteen carpets, (F)  twelve foams, (P)  fifteen plastics and papers, and (T) seventeen textiles and fabrics. }
	%The database comprises accelerations (movement, tapping), sound (movement, tapping), surface photographs (with and without flash), surface infra-red reflectance scans, and binary metal detection readings for eleven subjects. 
	%The dataset comprises 2160 surface trials of eleven users for each surface.} % 10 trials from 1 expert for training  and 10 test trials from ten users for the testing phase
	\label{fig:lmt108}
\end{figure*}

\subsection{Multi-source Classification} 

Accommodating data from heterogeneous information sources for a collective classification decision is not straightforward and can be tedious to implement. 
Due to the translation invariance of the kernel and our median heuristic choice of the hyperparameter $\sigma$, the MMD test automatically deals with many preprocessing steps that would otherwise be required. 
However, the MMD can deviate across information sources due to varying magnitudes in different dimensions and data streams.
Therefore, the method of unifying MMD scores from individual information sources to an overall classification decision is critical.

In addition to averaging MMDs from repeated tests for the same comparison, we separately compute the MMD within each information source and combine the source-specific values into a global score.
Based on the MMD estimator in Eq.~\eqref{eqn:mmdComputable}, we define the global discrepancy score
\begin{equation}
\label{eqn:geoMean}
% \mathrm{MMD}^{2}_b[\mathbb{P}_Y,\mathbb{P}_Z] =  \prod_{s=1}^{S} ({\mathrm{MMD}^{2}_b}^s[\mathbb{P}_Y,\mathbb{P}_Z])^{w_s}
 \mathrm{DS}[Y^{q},Z^{q}] = \prod_{s=1}^{S} \left(\mathrm{MMD}^{2}_b[\mathbb{P}_{Y^{q}_s},\mathbb{P}_{Z^{q}_s}]\right)^{w_s}
\end{equation} 
between two trials, ${Y^q}$ and ${Z^q}$.
This formulation is analogous to a weighted geometric mean or the arithmetic mean of individual logarithm-transformed values.
This log-average operator fosters MMD scale-invariance across information sources.
All individual MMD scores within each information source can have a weight $w_s\in \mathbb{R}^+$ for the overall classification decision of an unlabeled surface.
While the weights $w_s$ allow one to consider some information sources more strongly than others, one may also choose unit weights for all MMD scores, as we show in the experiments.

\subsection{k-Nearest-Neighbor Classification Decision}

The discriminator of our recognition framework makes the classification prediction based on the concept of $k$-nearest neighbors ($k$-NN) from global DS scores in Eq.~\eqref{eqn:geoMean}.
We are able to predict the identity of the unlabeled surface by means of the highest probabilistic agreement with a surface in the library. 
The unlabeled surface trial $Z^{q}$ will be classified to the class $c$ in the library with $C$ surfaces according to
\begin{equation}
 \label{eqn_classDecision}
 \min_{c \in C} \mathrm{DS}[Y^{q}(c),Z^{q}];
\end{equation}
it matches to the trial pair for which the global DS is smallest, \ie the nearest neighbor ($1$-NN).
In general, the classification decision could also be inferred from more nearest neighbors, which would require $k$ to be chosen during training.

\section{Experiments}

To show the efficiency of our recognition framework, we benchmark our algorithm against state-of-the-art classifiers on the open surface dataset by Strese et al.\ \cite{strese2017content} from the Lehrstuhl für Medientechnik (LMT) of TU Munich, Germany.
We demonstrate that our statistical approach automatically captures the most salient surface properties in the data without the need to learn surface classes, outperforming other classifiers with expertly crafted features.

\subsection{Multimodal Surface Dataset}

% General
The LMT108 dataset includes multimodal measurements from 108 surface textures (Fig.~\ref{fig:lmt108}), which were acquired with a handheld sensorized stylus that has a stainless steel tool-tip \cite{strese2017content}.
Each of the $108$ surfaces belongs to one of the nine material categories: (M) meshes, (S) stones, (G) metals/glasses/ceramics, (W) woods, (R) rubbers, (C) carpets, (F) foams, (P) plastics/papers, and (T) textiles/fabrics.
The number of surfaces in each category varies from five to seventeen.
Each recorded surface trial consists of nine information sources: two images (with and without flash); sound and acceleration from tapping action; sound, acceleration, frictional forces, IR surface reflectance, and metal detection during dragging of the tool.
These information sources stem from six sensors: a camera (ca), a microphone (mi), accelerometer (ac), two force-sensing resistors (fr), an IR surface reflectance sensor (ir), and a metal detection sensor (me).
All data streams apart from the images are a function of time.
For each trial, the information sources from tool dragging cover $4.8$ seconds of data without abrupt transient movements.
The accelerations and images each have three channels (axes and colors, respectively), and the forces have two channels.
The remaining information sources are all one-dimensional.
Both expert and user sets include ten trials per surface ($2160$ in total), but they differ in the number of humans (one expert vs.\ ten users) who acquired the data.
The expert considered different motions, scan velocities, and forces to ensure intra-class variance for each surface. %intentionally
The ten other users moved the stylus freely without any prescribed motion.

\begin{figure}[t]
	\includegraphics[width=83mm]{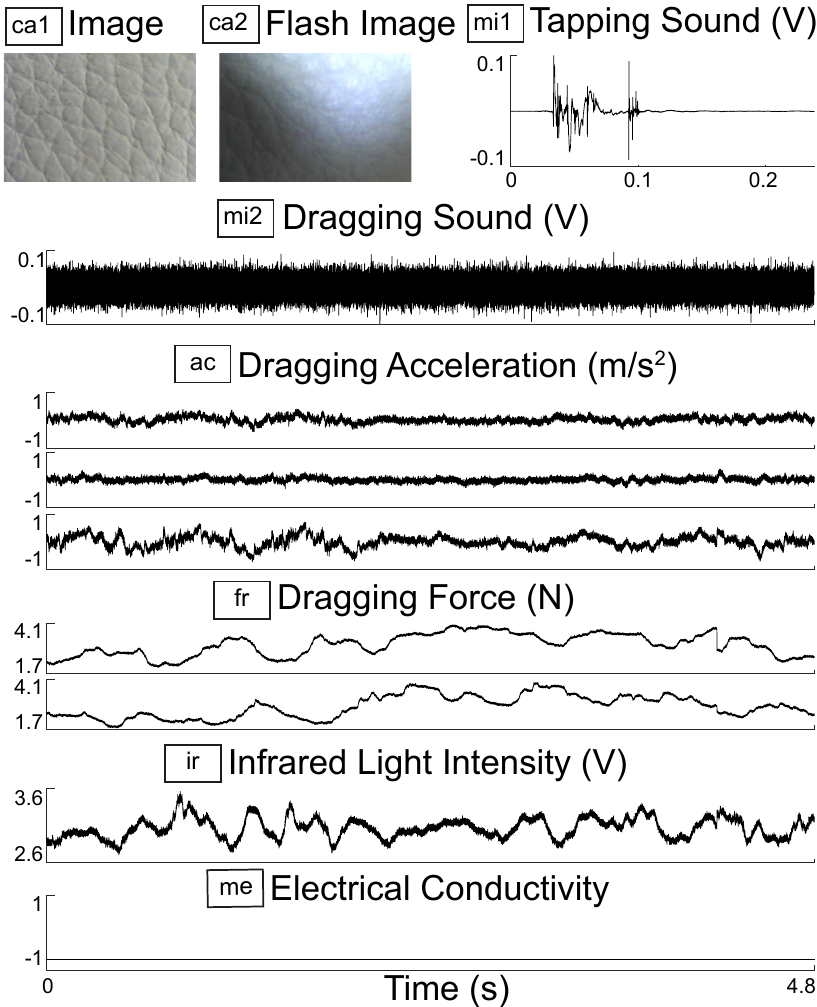}
	\centering
	\caption{The eight information sources of a user trial (testing phase) for a leather surface from the LMT108 dataset \cite{strese2017content}.
	The measurements originate from a camera (ca), a microphone (mi), an accelerometer (ac), two force-sensing resistors (fr), an IR reflectance sensor (ir), and a metal detection sensor (me).}
	\label{fig:leatherTrial}
\end{figure}

We note that we could not consider the acceleration taps from the LMT108 dataset because they are not available for the ten users; the online archive contains repeated data from the expert, and the original trials are lost.
Thus, we use only eight instead of the nine original information sources; see the sample data in Fig.~\ref{fig:leatherTrial}.
Acceleration transients from tapping  convey rich and descriptive information about the surface and would likely help the classification.
Qualitative comparisons are still possible despite this mismatch in the number of considered information sources.

\subsection{Data Preprocessing} 
Our recognition framework (Fig.~\ref{fig:pipeline}) uses all  sensor channels of the eight information sources for each LMT108 surface.
We represent the three-channel images in hue, saturation and value (HSV) color code.
The equidistant sampling scheme is used for data streams with a meaningful DC component, i.e., the images, forces, IR reflectance, and metal scans.
The remaining three information sources (mi1, mi2, ac) represent contact vibrations with distinct AC waveforms, and thus we perform sampling randomly in the frequency domain.
For these spectral auditory and haptic vibrations, we adopt the same frequency ranges (up to $7500$ Hz and $1000$ Hz respectively) in our algorithm as the authors of the LMT108 dataset \cite{strese2017content,strese2021haptic}.
Our criterion-based choice of the sampling domain (time vs.\ frequency) for time-series data matches the domains of the expertly crafted features in the original work \cite{strese2017content, strese2021haptic}.
Further, the cross-user compensation in Eq.~\eqref{eqn:scalShift} is applied to all information sources that are significantly affected by one or more of the following: scanning speed, applied force, tool dynamics, or session-dependent effects such as ambient noise.
For the LMT108 dataset, we apply the cross-user trick to five information sources: mi1, mi2, ac, fr, ir.

\subsection{MMD Test Setting} 

The number of data stream points varies across the eight information sources due to measurement duration, sampling rate, data dimensionality and time- or frequency-domain representation.
Regardless of the different sizes, we sample $n=m=400$ points from every data channel in each information source in the course of an individual MMD computation.
This extracted portion of data corresponds to approximately $0.3\%$ of each image, $0.8\%$ of the temporal force/reflectance/metal signals, $0.9\%$ of the dragging sound and $14.4\%$ of the tapping sound ($44.1$ kHz), and $6.1\%$ of the dragging acceleration ($10$ kHz).
We also repeatedly ($R=10$) perform the kernel two-sample test for all surface trial comparisons for each information source and average the $R$ MMD scores.
For the five information sources with the equidistant sampling scheme, the data points from the initial image area or time window (for the randomly selected first data point) do not exceed $1.15\%$ of the total number of corresponding data stream points.

To cover the data broadly with the prescribed number of data points $n$ and $m$ for each MMD test, the equidistant sampling strategy results in distances $T=11.8$ ms for the three time-series sources (fr, ir, me) and  $d_a=17$ pixels and $d_b=18$ pixels for both images ($480\times320$ pixels). %  15.9ms is for P=300
However, the sample time distance $T$ does not fulfill the mixing requirements for all force and IR reflectance surface signals; the metal signals mostly have one signal value and therefore are not suitable for calculating HSIC.
To fulfill the theoretical guarantees for vanishing autocorrelation times for all surfaces in both of these information sources, either a drastic increase in $T$ (decrease in the number of samples taken) or a measurement duration much longer than $4.8$ seconds would be required.
Longer time-series measurements would help to reduce inherent autocorrelations and most likely increase classification accuracy.
Here, we nonetheless stick to $400$ samples per channel to treat all information sources equivalently. 
Despite the mismatch of theoretically optimal and practically chosen $T$, the empirical results will show the efficacy of our approach.

The temporal metal detector reading is binary, \ie instantaneous detection of conductive material in the surface.
The LMT108 dataset contains twelve metallic surfaces. 
We would generally refrain from the use of the MMD test for binary data streams and simply focus on low-order statistical moments, e.g., the signal mean, as previously considered as a classification feature \cite{strese2017content}.
However, in order to treat all information sources equally for a consistent algorithm architecture, we apply the kernel two-sample test to the metal detection readings as well.
Therefore, we set the kernel length scale $\sigma=1$ whenever identical-value data would cause $\sigma=0$ with the heuristic in Eq.~\eqref{medianHeuristic}.
We further add a constant value of one to all MMD scores of the metal information source, so that $\mathrm{MMD}^{\mathrm{me}} \neq 0$ ensures no information losses with geometric means in Eq.~\eqref{eqn:geoMean}. 
We refrain from applying the cross-user compensation on the metal detection measurements because aligning the distribution mean would remove the information about material conductivity.

As mentioned before, the weighted geometric mean framework in Eq.~\eqref{eqn:geoMean} enables different influence strengths to be assigned to information sources in a single MMD score for the overall classification decision.
However, here we do not tune a model in training, and we thus simply weight all information sources equally, \ie $w_s=w=1$.

\subsection{Ablation Studies}

To demonstrate the functionality our proposed pipeline, we evaluate key aspects of our classifier through ablation studies.
We separately remove the following three components of our recognition pipeline (Fig.~\ref{fig:pipeline}): the RGB-to-HSV conversion for images, the DFT time-to-frequency-domain conversion for high-frequency vibrations, and the cross-user compensation for the five selected time-series sources.
Therefore, standard RGB images are considered in the HSV ablation study, and the vibration signals (mi1, mi2, ac) undergo an equidistant temporal instead of a spectral sampling procedure in the DFT ablation study.
Our reproducible code ensures sampling of the same data points with and without a given functionality, so that its pure effect can be observed. 
For deeper insight, we assess the classifier performance with full multimodal information as well as with individual information sources in the presence or absence of a functionality. 
In addition, we perform another train-test split by conducting surface recognition based on only the expert trials (1 expert vs.\ 1 expert) to primarily elucidate the effectiveness of the cross-user compensation.
For more insight, we report the performance in this only-expert-trials classification scenario for the other two ablation studies as well.
The 1 vs.\ 1 data partition results in a simpler classification task due to smaller signal variance from only one human and therefore (near-)perfect recognition rates with multimodal information.
Thus, we report the classifier performance only with single information sources.

\subsection{Comparison Methods}

The following briefly describes the six previously published classifiers to which we compare our approach. These all use expertly crafted features and are widely used for surface classification; see~\cite{flach2012machine,mohri2018foundations} for details. 
% Gaussian NB
Naive Bayes classifiers are a set of probabilistic classifiers that assume strong independence between features.
If the independence assumption holds, this algorithm performs well with less training data compared to other machine-learning models.
This classifier will choose the class with highest probability of joint likelihood between the given observation and the class occurrence.
Gaussian Naive Bayes models further assume that the joint likelihood follows a normal distribution.

% SVM
The SVM classifier is a discriminative model that learns a decision function from a training set to assign inputs to classes. %, supervised
The training phase can include a map to a higher-dimensional space, for instance through the kernel trick in Eq.~\eqref{kerneltrick},  to maximize the gap between two classes and accommodate a non-linear decision boundary.
A multiclass SVM uses multiple binary classifications to make an overall class prediction. The main difference to our approach is that single data points instead of whole distributions are mapped into RKHS.

% kNN
Given a data point in the test set, the non-parametric $k$-NN algorithm considers the feature distance to the $k$-nearest neighbors in the training set for a plurality vote to output a prediction. 
$k$-NN classifiers can differ in the  metric used (e.g., Euclidean or Mahalanobis distance) and in the weighting scheme for the nearest neighbors. 
The classical $k$-NN and our approach both make a prediction based on the nearest-neighbor concept between training and testing sets; however, $k$-NN considers distances between explicit features in contrast to our approach of quantifying distribution differences directly from the input data. % we do not compute features

% Ensemble
Ensemble classifiers adopt the basic concept of decision trees and combine multiple classifiers for the overall prediction, \ie differing classifiers or the same classifier with different configurations. 
This algorithm category considers a set of features and instances from the feature space and duplicates  or weights the instances for the final majority voting of individual classifiers.
This operation of altering the input feature space before training, also called bootstrap aggregating, fosters classifier robustness. 

% Random Forest
Random Forests fall into the category of ensemble methods and use bootstrapped feature spaces.
In this process, a reduced set of features is randomly selected for training each decision tree.
For a test instance, a majority voting scheme of all tree predictions determines the final decision.
As such classifiers require rigorous training, they are computationally expensive for large datasets.

% Adaboost % use citation?
Another popular ensemble classifier is adaptive boosting or AdaBoost;
it uses many decision trees that split on a single feature. 
In training, each new tree is constructed by taking the errors of the previous tree into account to reduce remaining misclassified observations.
In contrast to unit tree weights for random forests, AdaBoost trees with fewer errors will receive a higher final voting power, so that multiple weak classifiers can result in a better overall classifier.

\subsection{Performance Metrics}

Strese et al.~\cite{strese2021haptic,strese2017content} used $10$-fold cross validation (each fold: $1$ expert $+$ $9$ users vs. $1$ user) with the aforementioned baseline classifiers on the LMT108 dataset.
In contrast, our recognition pipeline considers only the expert trials (half of the LMT108 data)  for the library in the training phase, and thus addresses a substantially more difficult classification setting ($1$ expert vs.\ $10$ users).
The latest publication of Strese et al.~\cite{strese2021haptic} evaluated LMT108 classifier performance through the mean and standard deviation of the $10$-fold classification accura
In the previous publication \cite{strese2017content} also by Strese et al., a Euclidean-based $k$-NN classifier was evaluated by means of the classification precision.
We adopt the same two performance metrics for the LMT108 surface-recognition task to allow direct comparisons.

The classification accuracy for class $c$ is
\begin{equation}\label{eqn_acc}
\mathrm{Accuracy}(c) = \dfrac{ \mathrm{TP}(c)+\mathrm{TN}(c)}{\mathrm{TP}(c)+\mathrm{TN}(c)+\mathrm{FP}(c)+\mathrm{FN}(c)},
\end{equation}
with  true positives and negatives,  $\mathrm{TP}(c)$ and $\mathrm{TN}(c)$, and false positives and negatives, $\mathrm{FP}(c)$ and $\mathrm{FN}(c)$.
We compute the classification accuracy of ten user-specific folds (each test set has $108$ trials) and calculate their mean and standard deviation for the overall accuracy.
Further, the classification precision of a surface class $c$ is given by
\begin{equation}\label{eqn_prec}
\mathrm{Precision}(c) = \dfrac{\mathrm{TP}(c)}{\mathrm{TP}(c) + \mathrm{FP}(c)}.
\end{equation}
The class precision represents the ratio of correct predictions out of all predictions for class $c$.
We obtain the overall precision by averaging all class-specific values.

\subsection{Results and Discussion}

\begin{table}[b!]
\caption{Accuracy (mean $\pm$ standard deviation) and precision for expertly crafted classifiers compared to variations of our approach with $S$~information sources on the LMT108 dataset}
\resizebox{\columnwidth}{!}{%
\begin{tabular}{@{}llll@{}}
%\toprule
Algorithm & Accuracy [\%] &  $S$ & Sensor(s)\\ \midrule
Gaussian Naive Bayes \cite{strese2021haptic} & 78.1 $\pm$ 0.3  & 9 & 6\\
SVM (RBF) \cite{strese2021haptic} & 79.1 $\pm$ 0.5 &  9 & 6\\
k-NN (Mahalanobis) \cite{strese2021haptic} & 84.1 $\pm$ 0.4 & 9 & 6 \\
AdaBoost \cite{strese2021haptic} & 86.8 $\pm$ 0.8  & 9 & 6\\
Random Forests \cite{strese2021haptic} & 91.2 $\pm$ 0.8  & 9 & 6\\
Ours: visual (ca1+ca2) & 74.8 $\pm$ 2.5  & 2 & 1\\
Ours: auditory (mi1+mi2) & 54.8 $\pm$ 16.3 & 2 & 1\\
Ours: haptic (ac+fr) & 61.6 $\pm$ 6.0  & 2 & 2\\
%\textbf{Our Approach (V+)} & \textbf{77.3 $\pm$ 4.9}  & 2 & \textbf{3}\\
Ours: visual+auditory (ca1+ca2+mi1+mi2) & 84.1 $\pm$ 7.3  & 4 & 2\\
Ours: visual+haptic (ca1+ca2+ac+fr) & 92.7 $\pm$ 2.2 & 4 & 3 \\
Ours: auditory+haptic (mi1+mi2+ac+fr) & 74.2 $\pm$ 10.9 & 4 & 3 \\
\textbf{Ours: all (ca1+ca2+mi1+mi2+ac+fr+ir+me)} & \textbf{97.2 $\pm$ 2.1}  & \textbf{8} & \textbf{6}\\
Ours: all -- HSV color ablation & 95.8 $\pm$ 2.7 & 8 & 6 \\
Ours: all -- DFT ablation  & 95.1 $\pm$ 3.3 & 8 & 6 \\
Ours: all -- cross-user compensation ablation & 88.7 $\pm$ 6.7 & 8 & 6 \\
\midrule \midrule  
 & Precision [\%] &   & \\ 
 \midrule
k-NN (Euclidean) \cite{strese2017content} & 86 & 9 & 6\\ 
\textbf{Ours: all (ca1+ca2+mi1+mi2+ac+fr+ir+me)} & \textbf{98}  & \textbf{8} & \textbf{6}\\
%\bottomrule
\label{tab:results}
\end{tabular}
}
\end{table}

\begin{figure*}[t!]
	%	\centering
%	\includegraphics[width=170mm]{Figures/ClassResults_letter7_1.pdf}
	\includegraphics[width=170mm]{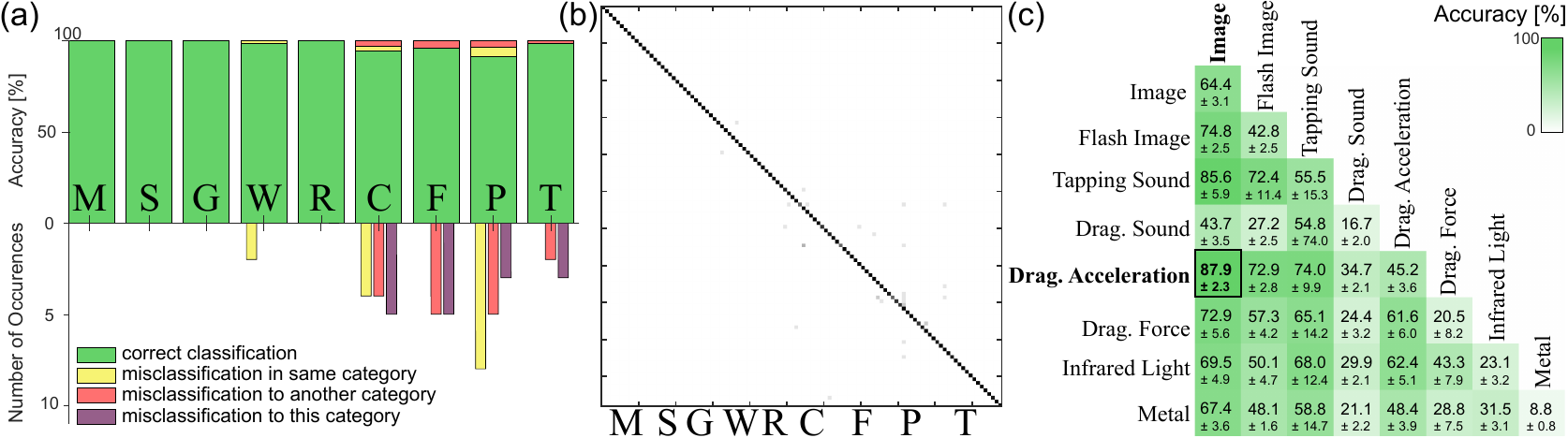}
	\centering
	\caption{Prediction results of our approach: (a) Relative and absolute statistics of the classification accuracy and (b) corresponding confusion matrix with all eight information sources ($S=8$) for $108$ surface classes ($C=108$). 
	For better illustration of similar texture types, we present the results of the $108$ surfaces grouped by the previously used material taxonomy.
	%Meshes, stones and glossy surfaces ($31$ textures) exhibit $100\%$ prediction accuracy, whereas categories such as carpets and papers/plastics experience a small number of inter- and cross-confusions. 
	(c) Classification accuracy for individual information sources and for their pairwise combinations. The surface image and contact-elicited dragging acceleration (boxed) provide complementary and robust sources for surface recognition. }%\commentbk{add true and prediction to (b)}}
	\label{fig:classResults}
\end{figure*}

As shown in Table~\ref{tab:results}, we achieve an accuracy of $97.2\%$ for $108$ surface classes with our full classifier, outperforming the next best model (Random Forests) by $6$\% despite using one less information source. 
We also achieve a mean classification precision of $98\%$ for all $108$ surfaces, which is $12\%$ higher than the Euclidean-based $k$-NN classifier \cite{strese2017content}. 
Thus, our classifier has very high positive predictive power. 
These two key results demonstrate that surface classification is more efficient with our automated approach of detecting maximal distribution differences than all previously developed classifiers that use features based on expert knowledge.

\paragraph*{Efficient surface classification with images} With regard to the unimodal surface-classification performance shown in Table~\ref{tab:results}, the visual modality (two three-channel images, ca1+ca2) is very efficient.
With only two images, we almost reach the classification accuracy of the multimodal Gaussian Naive Bayes classifier with all data streams.
Considering RGB instead of HSV images for the color ablation study results in slightly lower recognition accuracy ($95.8\% \pm 2.7\%$) compared to our full classifier, showing that our approach is only slightly harmed by this change. 
However, HSV images perform significantly better in single-image classification than their RBG representation in both train-test splits and with and without cross-user compensation (Table~\ref{tab:singleInfoAblation}).
We believe HSV matters more for single-modality classification because this color space encapsulates meaningful surface information in separate data channels. 
The decreased sensitivity to image lighting variations that HSV affords enables more robust image classification.
We further observe a superior performance of the standard HSV images (ca1) over the flash images (ca2) in both color spaces, both train-test splits, and with and without user compensation (Table~\ref{tab:singleInfoAblation}).
While the flash of the camera can reveal useful information regarding surface reflectivity, some descriptive properties of un-illuminated surface images seem to be lost for the recognition task.
Moreover, single-channel recognition with regular HSV images (ca1) shows that image hue ($42\%$) and saturation ($34\%$) are more informative than brightness value ($27\%$).
This effect is no longer significant with flash images (H: $19\%$, S: $18\%$, V: $17\%$) as the lightning conditions balance the descriptive information across the three image channels.

\paragraph*{Temporal and spectral time-series sampling} The DFT ablation study   
shows that our full classifier performs only slightly worse ($95.1\% \pm 3.3\%$) when using temporal sampling for the high-frequency vibration sources (Table~\ref{tab:results}).
The ablation studies with individual sources reveal that our preferred criterion-based domain for sampling tends to perform better for most information sources in both train-test splits and both user compensation settings (Table~\ref{tab:singleInfoAblation}). 
If we perform MMD tests on data from both the time- and frequency-domains for the same training data, we obtain a slightly higher recognition accuracy of $98.2\% \pm 1.7\%$.
Then, we have twelve instead of six time-series sources and perform the cross-user trick for the same five information sources in both the temporal and spectral domains. % 10 in total
As this performance improvement entails twice as many kernel evaluations for the time series, we conclude that a single criterion-based domain for sampling is already very efficient.

% 1 column table 
\begin{table}[b!]
  \centering
  \caption{Classification accuracy of  ablation studies for individual information sources for two train-test splits}
   \resizebox{\columnwidth}{!}{%
    \begin{tabular}{ccccc}
    %\toprule
    \multirow{2}[0]{*}{\shortstack{Info.\\ Source}} & \multirow{2}[0]{*}{ \shortstack{Train-Test\\ Split}} & \multirow{2}[0]{*}{\shortstack{Cross\\-user\\ Compens.}} & \multicolumn{2}{c}{Accuracy [\%]} \\
         &      &      & \shortstack{Temporal\\ Sampling}    & \shortstack{Spectral\\ Sampling} \\    
    %\multicolumn{1}{c}{Information Source} & \multicolumn{1}{c}{Train-Test Split} & Cross-user Compensation    & \multicolumn{1}{c}{Temporal Sampling} & \multicolumn{1}{c}{Spectral Sampling} \\
    \Xhline{2\arrayrulewidth} %\hline \midrule %\midrule
    \multirow{4}[0]{*}{mi1} & \multirow{2}[0]{*}{1 expert vs. 10 users} & no    & 50.9 $\pm$ 16.4    & \color{red} 48.4 $\pm$ 18.3 \\
         &      & yes    &  \color{red}47.6 $\pm$ 18.0    & \color{Green}\textit{55.9 $\pm$ 15.4} \\\cmidrule{2-5}
         & \multirow{2}[0]{*}{1 expert vs. 1 expert} & no    & 22.4 $\pm$ 5.0    & \color{red} 19.0 $\pm$ 3.1 \\
         &      & yes    &   21.1 $\pm$ 3.5    & \color{Green} 27.1 $\pm$ 3.7 \\ \Xhline{2\arrayrulewidth} %\hline
    \multirow{4}[0]{*}{mi2} & \multirow{2}[0]{*}{1 expert vs. 10 users} & no    & 4.2  $\pm$  1.2 	
    & \color{red} 6.6  $\pm$  1.3	 \\
         &      & yes    & \color{red} 4.3  $\pm$  1.5   & \color{Green}\textit{15.3  $\pm$  3.3}	 \\\cmidrule{2-5}
         & \multirow{2}[0]{*}{1 expert vs. 1 expert} & no    & 12.6  $\pm$  2.9	    & 39.9  $\pm$  4.3 	 \\
         &      & yes    & 13.0  $\pm$  2.6 	   & 41.0  $\pm$  2.7 	 \\ \Xhline{2\arrayrulewidth} %\hline
    \multirow{4}[0]{*}{ac} & \multirow{2}[0]{*}{1 expert vs. 10 users} & no    & 48.3  $\pm$  5.6	    & \color{red} 26.4  $\pm$  1.8	
	 \\
         &      & yes    & 45.8  $\pm$  4.5	    & \color{Green}\textit{43.8  $\pm$  2.3}  \\\cmidrule{2-5}
         & \multirow{2}[0]{*}{1 expert vs. 1 expert} & no    & 46.9  $\pm$  9.8	    & \color{Green} 72.8  $\pm$  6.9	 \\
         &      & yes    & 52.6  $\pm$  7.5	   & \color{red} 56.6  $\pm$  7.5	 \\ \Xhline{2\arrayrulewidth} %\hline\hline
    \multirow{4}[0]{*}{fr} & \multirow{2}[0]{*}{1 expert vs. 10 users} & no    & \color{red} 14.2  $\pm$  3.9	    & 5.6  $\pm$  1.7	 \\
         &      & yes    & \color{Green}\textit{20.2  $\pm$  7.5}   & \color{red}7.1  $\pm$  2.2 	 \\\cmidrule{2-5}
         & \multirow{2}[0]{*}{1 expert vs. 1 expert} & no    & \color{Green} 33.5  $\pm$  12.2    & 22.0  $\pm$  6.3 	 \\
         &      & yes    & \color{red} 12.4  $\pm$  4.8	   & 10.6  $\pm$  4.1 	 \\ \Xhline{2\arrayrulewidth} %\hline
    \multirow{4}[0]{*}{ir} & \multirow{2}[0]{*}{1 expert vs. 10 users} & no    & 23.7  $\pm$  4.6   & 8.0  $\pm$  2.1 	 \\
         &      & yes    & \color{Green}\textit{23.5  $\pm$  3.7}	    & \color{red}12.3  $\pm$  2.8	 \\\cmidrule{2-5}
         & \multirow{2}[0]{*}{1 expert vs. 1 expert} & no    & \color{Green}55.5  $\pm$  5.1	    & 28.0  $\pm$  4.3	 \\
         &      & yes    & \color{Red} 23.8  $\pm$  4.5	   & 24.9  $\pm$  2.7	 \\   \Xhline{2\arrayrulewidth} %\hline   
    \multirow{4}[0]{*}{me} & \multirow{2}[0]{*}{1 expert vs. 10 users} & no    &  \color{Green}\textit{8.5  $\pm$  1.3}   & \color{red} 8.1  $\pm$  1.2 \\
         &      & yes    & 7.6 $\pm$ 1.3    & 4.4 $\pm$ 1.5 \\\cmidrule{2-5}
         & \multirow{2}[0]{*}{1 expert vs. 1 expert} & no    & 8.8 $\pm$ 0.8    & 9.4 $\pm$ 0.9 \\
         &      & yes    & 7.9 $\pm$ 0.8   & 5.4 $\pm$ 1.2 \\              
\midrule \midrule  
    \multicolumn{1}{c}{ } & \multicolumn{1}{c}{ } &      & \multicolumn{1}{c}{RGB} & \multicolumn{1}{c}{HSV} \\ \Xhline{2\arrayrulewidth} %\hline\midrule %\midrule  
    \multirow{4}[0]{*}{ca1} & \multirow{2}[0]{*}{1 expert vs. 10 users} & no    & \color{red} 53.3 $\pm$ 2.6    & \color{Green}\textit{65.1 $\pm$ 2.9} \\
    % (spatial) in caption
         &      & yes  & 49.0 $\pm$ 2.6  & 64.1 $\pm$ 3.6 \\\cmidrule{2-5}
         & \multirow{2}[0]{*}{1 expert vs. 1 expert} & no     &   76.7 $\pm$ 4.8  & 84.6 $\pm$ 4.9 \\
         &      & yes     & 68.3 $\pm$ 3.9   & 82.3 $\pm$ 4.1 \\  \Xhline{2\arrayrulewidth} %\hline    
    \multirow{4}[0]{*}{ca2} & \multirow{2}[0]{*}{1 expert vs. 10 users} & no     & \color{red} 39.1 $\pm$ 2.0    & \color{Green}\textit{42.6 $\pm$ 1.7} \\
         &      & yes     & 35.4 $\pm$ 3.6  & 43.5 $\pm$ 3.6 \\\cmidrule{2-5}
         & \multirow{2}[0]{*}{1 expert vs. 1 expert} & no    &  65.5 $\pm$ 5.8  & 74.2 $\pm$ 5.7 \\
         &      & yes     & 61.4 $\pm$ 4.9   & 70.4 $\pm$ 5.3 \\                  
    %\bottomrule
    \end{tabular}%
  \label{tab:singleInfoAblation}%
 }
\end{table}

% User variations exists
\paragraph*{Sources of cross-user fluctuations} With regard to the effect of the user on classification accuracy, our algorithm has a larger variability of user-specific mean accuracies than the previously published machine-learning classifiers.
In our optimal setting ($S=8$), the accuracy of the testing set of each user ranges from 92\% to 99\% (for seven users $\geq 97$\%). 
This larger variance stems from our choice to include  only the expert data instead of additional data from the other nine users for the library in cross-validation.
These fluctuations originate from session-dependent (e.g, ambient noise) and user-dependent (e.g., motion and tapping strength) effects that diversify the surface data, so that prediction is more difficult for unseen data from training with fewer humans.
The haptic and auditory modalities exhibit particularly high user variance.
The larger variance of auditory over haptic information sources may arise from the fact that audio signals are more prone to ambient noise than haptic data, which can also have characteristic low-frequency components. 
We confirm our hypothesis about the source of fluctuations by adopting the same train-test split (1 expert + 9 users vs.\ 1 user) as the baseline with their cross-validation approach (no cross-user trick).
This simpler classification problem results in a very high accuracy ($99.1\%\pm 0.9\%$) with similar user variance compared to the baselines.
This remains a cross-user train-test setting, and adding our cross-user compensation to the same five information sources increases the performance ($99.4\% \pm 0.8\%$).
In summary, classification performance and robustness can benefit from a larger library with data from more users, albeit at a higher computational cost.
Nevertheless, these results also confirm that our cross-user compensation strategy is very efficient with a data library gathered by only one expert. 

\paragraph*{Efficient cross-user compensation} The cross-user-compensation ablation study (Table~\ref{tab:results}) shows that removing our cross-user trick reduces the mean accuracy of our proposed full classifier by almost $9\%$ (to $88.7\%$) and also increases the standard deviation across users by 4.6\% (to $6.7\%$).
In general, kernel two-sample testing benefits from using all statistical moments of the two distributions to compare. 
However, aligning the mean of the distributions significantly boosts surface-recognition performance and make it more robust across users. 
We believe this cross-user trick works well because session- and user-dependent effects are concentrated in the mean of the extracted distributions, whereas the higher-order statistical moments mainly convey discriminative surface information.
From the ablation studies with individual information sources (Table~\ref{tab:singleInfoAblation}), we report better performance outcomes for all vibrations when incorporating the cross-user trick with spectral sampling in the 1 vs.\ 10 setting.
The effect of the cross-user trick on the vibration sampling in the temporal domain is less striking.
In contrast, classification performance with temporal sampling of frictional forces (fr) benefits by $6\%$ when the cross-user trick is applied.

For the two images, applying the cross-user trick does not greatly degrade recognition performance.
This might be because the smartphone camera system used to capture these images automatically reduces certain user- and session-dependent image effects in hardware and software.
Nevertheless, we obtain a high classification accuracy of $97.1\% \pm 2.2\%$ when we apply the cross-user trick to the two images as well as the five time-series sources. 
While the proposed compensation trick is mainly intended for cross-user classification, using the trick in expert-trial classification (1 vs.\ 1) drastically reduces the recognition rate for the three information sources of ac (spectral), fr, and ir (Table~\ref{tab:singleInfoAblation}).
Therefore, meaningful information about surface identity is lost in the course of the alignment of these distribution means.
In contrast, we observe an improvement in the expert-trial classification for two information sources of mi1 and ac (temporal) when applying the cross-user trick. 
The tapping sound (mi1) is the only source for which expert-trial classification has a much lower recognition rate than the 1 vs.\ 10 train-test split.
This suggests that the ten tapping trials of the expert greatly differ, which confuses our single-source classifier in the 1 vs.\ 1 setting.
At the same time, this diverse tapping set significantly helps other information sources in cross-user recognition, as shown in Fig.~\ref{fig:classResults}(c).

% missing taps from user
\paragraph*{Missing acceleration tap data} We could not consider the missing acceleration taps for our classifier, which prevents a perfect comparison with the performance of previously published methods. 
Tapping is an exploratory procedure that greatly aids surface identification for humans~\cite{LaMotte00-JN-Softness}.
In dual-source classification, the tapping sounds, with the aforementioned diverse expert trials, lift the recognition accuracy of the other source by $21\%$ to $50\%$ (Fig.~\ref{fig:classResults}(c)).
As all the tapping sounds and accelerations are from the same training or testing interactions, 
our classifier would likely benefit from the missing acceleration source in the testing phase.
This benefit is expected because the additional acceleration data represents new transient information of the tool-surface interaction and because the MMD efficiently discriminates (new) data differences.
Even if the lost acceleration taps would degrade the performance of our multimodal classifier, its transparent architecture would effortlessly reveal this trend; in contrast to several machine- and deep-learning models, no retraining with another set of information sources would be necessary.

\paragraph*{Binary metal signals} We apply the kernel two-sample test to the binary metal recordings to have a uniform simple algorithm architecture for all information sources.
Considering only one of two subgroups (conductive material or not) in the data library based on the given metal detection reading would facilitate the recognition task and reduce the computational cost of our algorithm. 
Further, we shifted all MMD scores of this information source to avoid $\mathrm{MMD}=0$ in the geometric mean framework.
Shifting all MMD scores to have positive values also matters when not all decision power should be given to a single information source, for examples in safety-critical applications.

\paragraph*{Texture-specific insights} We can further investigate the performance of our optimal classification setting ($S=8$).
We  classify all $360$ test instances of the four harder surface categories (meshes, stones, glossy materials and rubbers) correctly, and
only (2/130) intra-category confusions occur for wooden materials (Fig.~\ref{fig:classResults}(a) and (b)).
Such hard surface textures amplify contact-elicited signals and therefore convey richer, more distinctive information.
Of our few critical misclassifications ($30/1080$), ten surface interactions are mistaken as a carpet or foam material.
The correct surface was among the top candidates in most of these misclassifications, hinting that one could also consider the $k>1$ nearest neighbors.
Glitter paper is a frequent false positive candidate ($8$ times); the abrupt stick-slip motions on such a texture may cause atypical irregularly spread contact signals that confuse the classifier.

\paragraph*{Complementary surface information} The classification performance of pairwise information sources (Fig.~\ref{fig:classResults}(c)) illustrates the superior classification accuracy ($87.9\%$) of an image together with dragging acceleration signals.
Intuitively, contact and non-contact sensors may provide complementary information about the surface. 
If these two sources (ca1+ac) turn out to be expressive in a recognition pipeline in the long run, the source weighting for the overall decision can be adjusted easily. 
For instance, we can reach the mean accuracy of $98.5\% \pm 1.2\%$ for the LMT108 surface set by increasing the weights ($w_{ca1}=w_{ac}=3$) of only these two information sources through Eq.~\eqref{eqn:geoMean}.

\paragraph*{Difficulty of multi-user surface classification} To the best of our knowledge, there exists no deep-learning algorithm trained for surface classification on the whole LMT108 dataset or any other multi-user surface dataset.
The existing references \cite{zheng2016deep,joolee2022deep,strese2021haptic} all consider data from only one human, and therefore they cannot be compared with our multi-user surface-recognition task, which we believe is more relevant for practical use.
In this context, we show through our ablation studies how a simple shift of a distribution mean can effectively compensate for spurious effects in time-series data and increase performance in a setting with limited training data captured by one human.

To get a sense of how our classifier performs compared to the most similar deep-learning references, we can see that our image plus tactile (ca1+ac) classifier achieves very high performance ($97.1\% \pm 1.2\%$) in the 1 vs.\ 1 setting on the LMT108 dataset; this performance is comparable to a deep-learning classifier \cite{zheng2016deep} with image and acceleration input from $69$ surfaces captured by one expert (part of the LMT69 dataset).
Note that we did not apply cross-user compensation to obtain these results given the findings of the relevant ablation study. % as they are 1vs1
Reporting multimodal surface-recognition performance with an extensive training set (e.g., through standard cross validation) may also facilitate future comparisons with deep-learning networks.
Thus, we report an accuracy of $88.9\% \pm 3.7\%$ with only standard images and $96.9\% \pm 1.4$\% with image plus tactile (ca1+ac) input for the 10 vs.\ 1 setting of the LMT108 dataset.
In this multi-user setting, we applied the cross-user trick only to the dragging acceleration signals.
Overall, our approach requires less data and exhibits higher interpretability as well as much simpler training efforts compared to deep-learning classifiers.

% Gaussian Kernel is all-rounder
\paragraph*{Kernel choice} The squared exponential kernel is a popular choice with well-established hyperparameter heuristics that beneficially do not require further kernel tuning.
This kernel function performs well on the LMT108 surface images and time series.
We did not systematically investigate other kernels but tried a few other choices early in our investigations.
For instance, the Laplace kernel together with the median heuristics seemed to result in similar recognition performance.
However, some other kernels require non-trivial hyperparameter optimization, which we want to avoid here.

%\IEEEpubidadjcol
\section{Conclusion}

Automated solutions are gaining popularity in big-data initiatives and are particularly relevant for emerging fields such as surface and object recognition.
Machine learning and deep learning have been extensively used for classification tasks, but they require expert knowledge, heuristics, and/or feature engineering. 
To circumvent these processes, we have developed an automated classification framework that examines images together with time series using the kernel two-sample test.
This easily implementable approach is based on the principle of comparing data in the space of probability distributions and automatically quantifying descriptive differences.
Our comprehensible, coherent algorithm architecture unlocks an elegant way to identify what sensor information improves classification performance.
We demonstrate for the task of surface recognition with $108$ classes that our approach achieves $6$\% higher accuracy than competitive multimodal baseline classifiers with expertly engineered features.
We achieve this performance with fewer sensor measurements, less training data, and little data processing in our standard algorithm setting. 
We also discuss how to further boost the recognition performance of our multimodal classifier, which practitioners can adopt and tune (e.g., kernel, source weights) for their use case.
New surface-interaction data captured with a more precise sensing system or with additional measurements (e.g., tool speed, tool angle) could provide additional insights for surface classification.
Compact, representative intra-class sets in the training library would leverage scalability.
Future work could also include applying modified versions of the MMD metric to improve test power and computational cost.
In addition, it would be of great interest to obtain comparable performance with fewer information sources and sparse training data to contrast with the requirements of existing deep-learning networks; this goal may necessitate more effort in the training phase.
We also envision using this approach for other learning tasks such as clustering.

		\bibliographystyle{IEEEtran}
		\bibliography{main_arXiv}

			\appendices 

\section{Software Implementations}
	\label{sec_softwareImplements}
We test the LMT108 surface dataset \textbf{(1)} with our open-source classification pipeline \textbf{(4)}, using unmodified implementations of the standard kernel two-sample test \textbf{(2)} and HSIC \textbf{(3)}.
For the reproducibility of our results, our open-source code includes a default setting for the random number generator.
The links to the original sources are as follows:\\
\textbf{(1)}: \url{https://zeus.lmt.ei.tum.de/downloads/texture} \\
\textbf{(2)}: \url{http://www.gatsby.ucl.ac.uk/~gretton/mmd/mmd.htm} \\
\textbf{(3)}: \url{http://people.kyb.tuebingen.mpg.de/arthur/indep.htm} \\
\textbf{(4)}: Publication of code after paper publication

\section{Mixing Properties of Surface Signals} 
		\label{sec_appendixMIXproperties}

The vanishing autocorrelation time $T^*$ ensures almost independent time-series samples for the kernel two-sample test. Additionally, it also has a physical interpretation and yields insight into certain system properties.

\begin{figure}[t]
	%	\centeringhttps://sharelatex.gwdg.de/project/6122d1fba5d6ed0094864ae3
	\includegraphics[width=83mm]{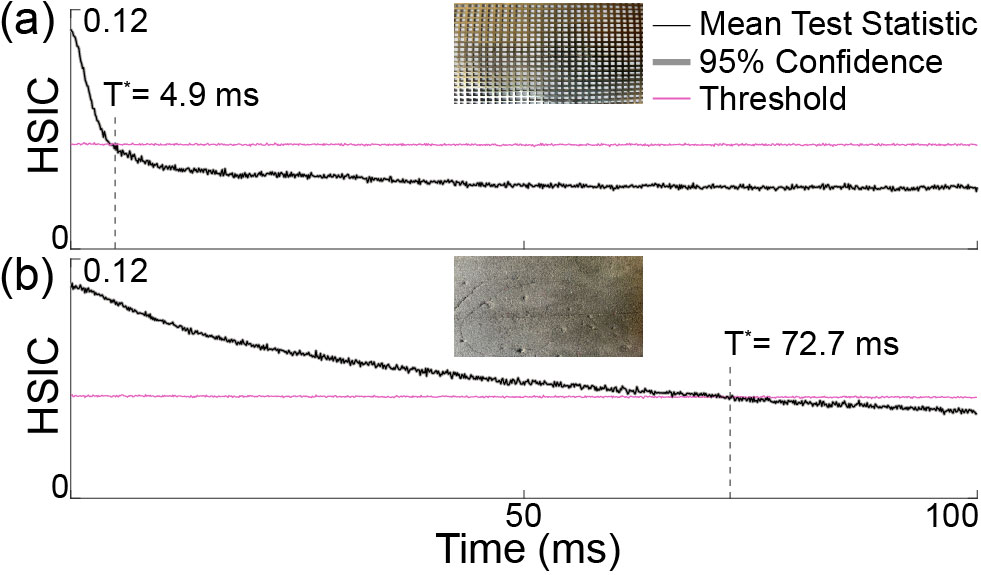}
	\centering
	\caption{HSIC ($R=500$, $\alpha=0.05$, $t_{1,\max}= 1$ s) as a function of time for measured IR light reflectance  on (a) an aluminum mesh and (b) EPDM foam (from LMT108 training data \cite{strese2017content}). 
	The mixing of IR sensor samples occurs roughly $15$ times faster for the metal mesh compared to the flat absorbing foam surface.}
	%(\textbf{C}) The vanishing auto-correlations times of reflectance scans and force readings fluctuates across surfaces and also varies within surface categories. 
	%\commentbk{move to appendix? last sentence could be also in text}}
	\label{fig:hsic}
\end{figure}

In the case of surface signals, these mixing properties express how rich different trials from interactions with a particular 
surface $c$ are for a given information source $s$.
For example, time-series measurements from an IR reflectance sensor provide interesting HSIC results on two different surfaces (Fig.~\ref{fig:hsic}).
While the aluminum mesh surface requires only about $5$ ms between samples for new uncorrelated information, the mixing for the foam surface is roughly $15$ times slower.
The three-dimensional porous mesh surface yields a rich signal, while the flat black foam surface greatly absorbs the IR signal and exhibits low signal richness.

On a broader scale, these surface-specific mixing properties provide a time-efficient guideline for acquiring new measurements for a given information source.
The measurement time can be individually adjusted for each surface to avoid autocorrelations and obtain rich independent data.

		\section*{Acknowledgments}

The authors thank Dr.\ Strese and Prof.\ Steinbach for their helpful description of the LMT database, Ben Richardson for helpful feedback on the manuscript, and the International Max Planck Research School for Intelligent Systems (IMPRS-IS) for supporting Behnam Khojasteh and Friedrich Solowjow.

		\ifCLASSOPTIONcaptionsoff
		\newpage
		\fi
			
	\begin{IEEEbiography}[{\includegraphics[width=1in,height=1.25in,clip,keepaspectratio]{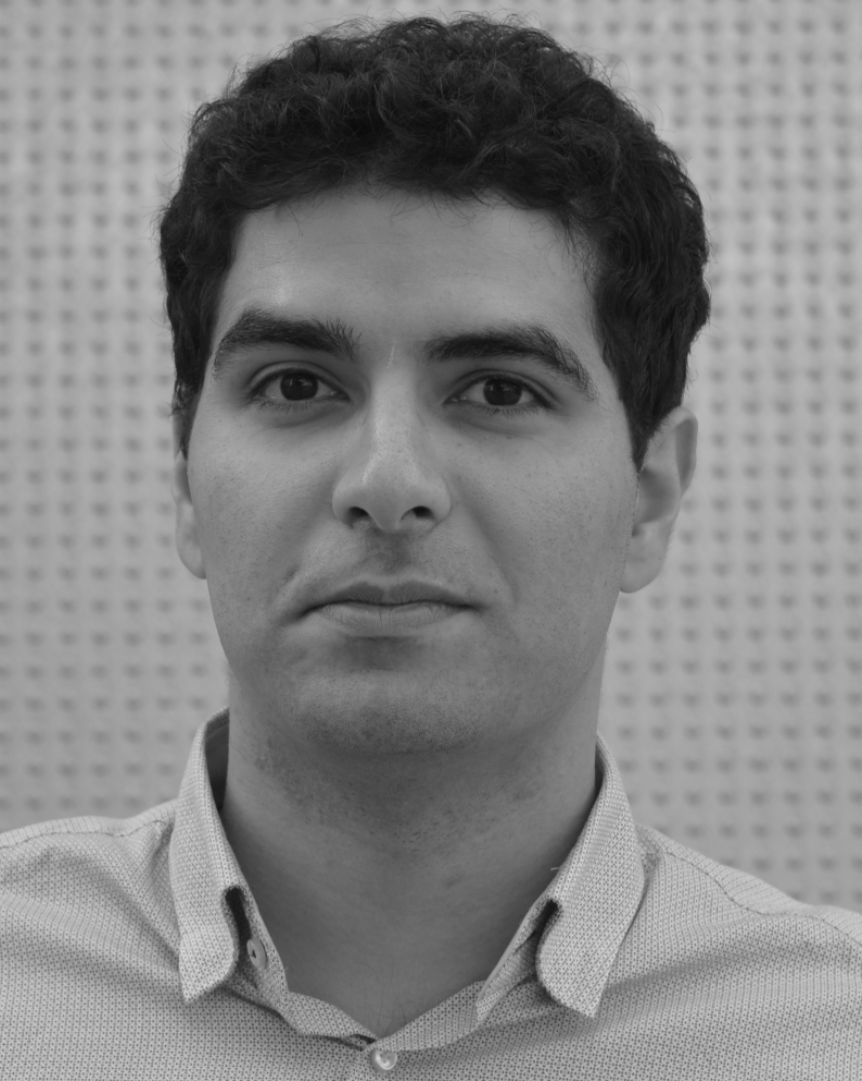}}]
	{Behnam Khojasteh}
	received the B.\,Sc.\ and M.\,Sc.\ degrees in Mechanical Engineering and Mechatronics in 2014 and 2017, respectively, from Hamburg University of Technology, Hamburg, Germany. 
	During his Master studies, he was a visiting researcher in the Department of Electrical and Computer Engineering at the University of California, Santa Barbara, USA.
	Khojasteh is currently a Ph.D. student in the Haptic Intelligence Department at the Max Planck Institute for Intelligent Systems, Stuttgart, Germany and a member of the International Max Planck Research School for Intelligent Systems.
	He has more than two years of industry experience in the automotive and aerospace field. 
\end{IEEEbiography}

\begin{IEEEbiography}[{\includegraphics[width=1in,height=1.25in,clip,keepaspectratio]{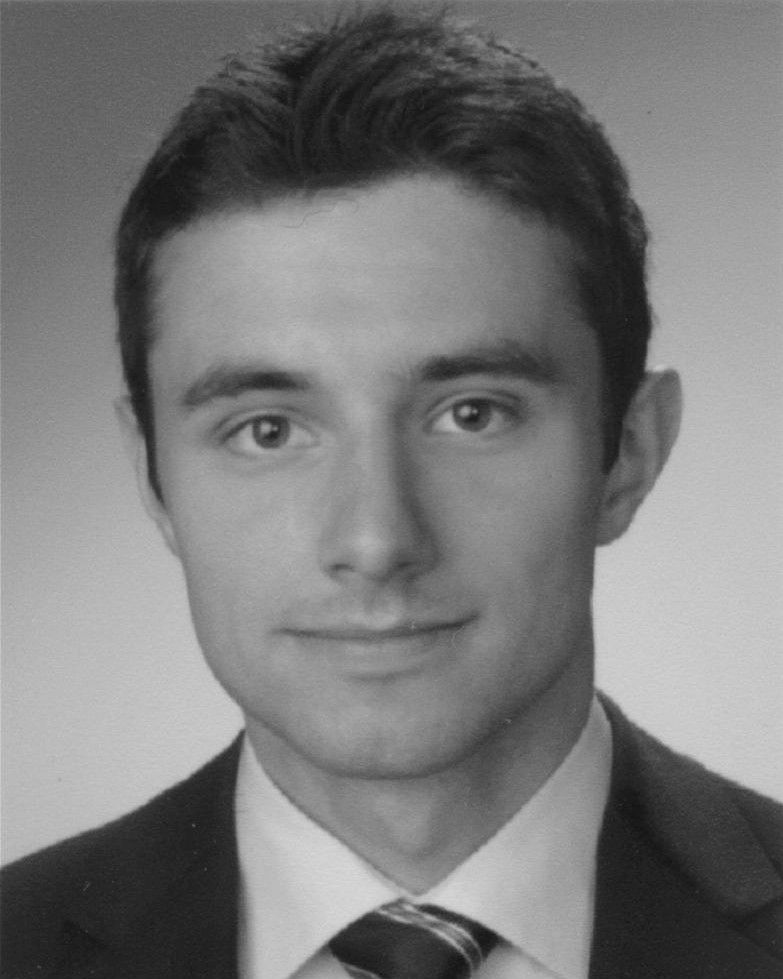}}]
{Friedrich Solowjow} received B.\,Sc.\ degrees in Mathematics and Economics from the University of Bonn in 2014 and 2015, respectively, and an M.\,Sc.\ degree in Mathematics also from the University of Bonn in 2017. 
In 2022, he received his Ph.\,D.\ (Dr.\,rer.\,nat.) from the University of Tübingen through the International Max Planck Research School for Intelligent Systems (IMPRS-IS).
Currently, he is a senior researcher (Akademischer Rat) in the Institute for Data Science in Mechanical Engineering at the RWTH Aachen University, Germany.
His main research interests are in the intersection between systems and control theory and machine learning.
\end{IEEEbiography}

%\newpage

\begin{IEEEbiography}[{\includegraphics[width=1in,height=1.25in,clip,keepaspectratio]{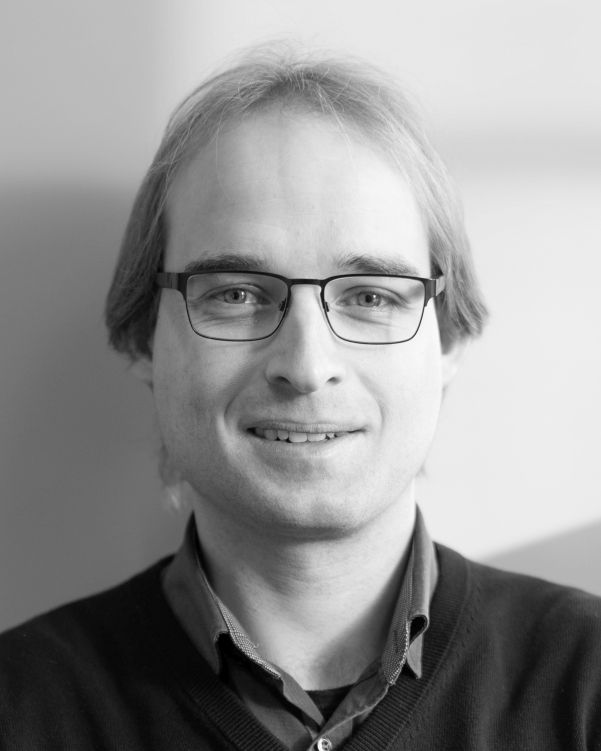}}]
{Sebastian Trimpe} received the B.\,Sc.\ degree in general engineering science and the M.\,Sc.\ degree (Dipl.-Ing.) in electrical engineering from Hamburg University of Technology, Hamburg, Germany, in 2005 and 2007, respectively, and the Ph.\,D.\ degree (Dr.\,sc.) in mechanical engineering from ETH Zurich, Zurich, Switzerland, in 2013.  Since 2020, he is a full professor at RWTH Aachen University, Germany, where he heads the Institute for Data Science in Mechanical Engineering.  Before, he was an independent Research Group Leader at the Max Planck Institute for Intelligent Systems in Stuttgart and Tübingen, Germany.  His main research interests are in systems and control theory, machine-learning, networked systems and robotics.  Dr.~Trimpe has received several awards, including the triennial IFAC World Congress Interactive Paper Prize (2011), the Klaus Tschira Award for achievements in public understanding of science (2014), the Best Paper Award of the International Conference on Cyber-Physical Systems (2019), and the Future Prize by the Ewald Marquardt Stiftung for innovations in control engineering (2020).
\end{IEEEbiography}

\begin{IEEEbiography}[{\includegraphics[width=1in,height=1.25in,clip,keepaspectratio]{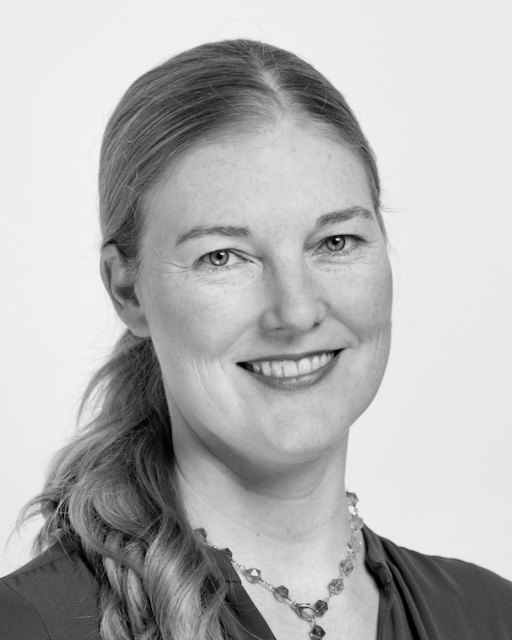}}]{Katherine J. Kuchenbecker} earned B.S., M.S., and Ph.D. degrees in Mechanical Engineering from Stanford University, USA, in 2000, 2002, and 2006, respectively. She did postdoctoral research at the Johns Hopkins University and was a faculty member in the GRASP Lab at the University of Pennsylvania from 2007 to 2016. Since 2017 she has been a director at the Max Planck Institute for Intelligent Systems in Stuttgart, Germany. Her research blends robotics and human-computer interaction with foci in haptics, teleoperation, physical human-robot interaction, tactile sensing, and medical applications. She delivered a TEDYouth talk on haptics in 2012 and has been honored with a 2009 NSF CAREER Award, the 2012 IEEE RAS Academic Early Career Award, a 2014 Penn Lindback Award for Distinguished Teaching, elevation to IEEE Fellow in 2022, and various best paper, poster, demonstration, and reviewer awards. She co-chaired the IEEE RAS Technical Committee on Haptics from 2015 to 2017 and the IEEE Haptics Symposium in 2016 and 2018.
\end{IEEEbiography}

		
		%\vfill
		


	\end{document}